\documentclass[conference]{IEEEtran}
\usepackage{hyperref} 
\usepackage{booktabs} 
\usepackage{microtype}
\usepackage{graphicx}
\usepackage{enumitem}
\usepackage{epsfig}
\usepackage{calc}
\usepackage{amssymb}
\usepackage{amstext}
\usepackage{amsmath}
\usepackage{amsthm}
\usepackage{multicol}
\usepackage{pslatex}
\usepackage{subcaption} 

\title{Style Transfer and Extraction for the Handwritten Letters Using Deep Learning}

\author{\IEEEauthorblockN{Omar Mohammed}
\IEEEauthorblockA{
Univ. Grenoble-Alpes\\
GIPSA-lab and LIG-lab\\
38000 Grenoble, France\\
omar-samir.mohammed@grenoble-inp.fr}
\and
\IEEEauthorblockN{Gerard Bailly}
\IEEEauthorblockA{
Univ. Grenoble-Alpes\\
GIPSA-lab\\
38000 Grenoble, France\\
gerard.bailly@grenoble-inp.fr}
\and
\IEEEauthorblockN{Damien Pellier}
\IEEEauthorblockA{
Univ. Grenoble-Alpes\\
LIG-lab\\
38000 Grenoble, France\\
damien.pellier@univ-grenoble-alpes.fr}
}

\date{}

\begin{document}
\maketitle 
\section*{Abstract}

\par How can we learn, transfer and extract handwriting styles using deep neural networks? This paper explores these questions using a \textit{deep conditioned autoencoder} on the IRON-OFF handwriting data-set. We perform three experiments that systematically explore the quality of our style extraction procedure. First, We compare our model to handwriting benchmarks using multidimensional performance metrics. Second, we explore the quality of style transfer, i.e. how the model performs on new, unseen writers. In both experiments, we improve the metrics of state of the art methods by a large margin. Lastly, we analyze the latent space of our model, and we see that it separates consistently writing styles.

\section{Introduction} \label{sec:introduction}

\par One aspect of a successful human-machine interface (e.g. human-robot interaction, chatbots, speech, handwriting~\ldots) is the ability to have a personalized interaction. This affects the overall human experience, and allow for a more fluent interaction. At the moment, there is a lot of work that uses machine learning in order to learn to model for such interactions. However, most of these models do not address the issue of personalized behavior: they try to average over the different examples from different people in the training set. Identifying the human styles during the training and inference time open the possibility of biasing the models output to take into account the human preference. In this paper, we focus the problem of styles in the context of handwriting.

\par However, defining and extracting handwriting styles is a challenging problem, since there is no formal definition for these styles (i.e. it is an ill-posed problem). A style is both social -- depends on writer's training, especially at middle school -- and idiosyncratic -- depends on the writer's shaping (letter roundness, sharpness, size, slope~\ldots) and force distribution across time. To add to the problem, till recently, there were no metrics to assess the quality of handwriting generation.  

\par There are two questions: what is the task itself? and what is the style used to achieve this task?. In handwriting, the task space is well defined (i.e. which letter we want to write), thus, allowing us to focus on the second part, of extracting styles for achieving this task.

In this paper, we address the problem of style extraction by using an conditioned-temporal deep autoencoder model. The conditioning is on the letter identity. The reason we use an autoencoder is that there is no explicit way that we know about to evaluate the quality of the handwriting styles other than using them to generate handwriting, and evaluate this generation. \cite{mohammed2018DTL} introduced benchmarks and evaluation metrics  in order to assess the quality of generating handwritten letters. In comparison to the those benchmarks and metrics, we achieve higher performance, while extracting a meaningful latent space. 

\par We also hypothesize that the latent space of styles is generic, i.e. that it will generalize over unseen writers, thus achieving a ``transfer of style''. To test this hypothesis, we assess our model on 30 new writers. We compare the tracings generated by this model to a benchmark model already proposed for online handwriting generation.

\par In addition, we explore the latent space of our model for each letter separately. This revealed  that there is a limited number of 'unique' styles per letter, categorical as well as continuous. We report our analysis for some of the letters, since a full analysis is out of the scope for this paper.

\par Thus, our contributions in this paper are the following:
\begin{itemize}
    \item We test and compare our deep conditioned autoencoder  with the state of the art benchmarks. We show that this model greatly improves the generation performance over a state of the art benchmark model.
    \item We experiment on performing style transfer on new writers using this model achieves, and we show that it achieves much better results than the benchmark model.
    \item Finally, and maybe most interestingly, we further analyze the extracted the latent space from our model to show that there is a limited number of styles for each letter and that the style manifold is not a continuous space.
\end{itemize}

\section{Related work}

\subsection{Generative models}
\par Recent advances in deep learning~\cite{Goodfellow-et-al-2016} architectures and optimization methods led to remarkable results in the area of generative models. 
For static data, like images, the mainstream research builds on the advances in \textit{Variational Autoencoders}~\cite{kingma2013auto} and \textit{Generative Adversarial Networks}~\cite{goodfellow2014generative}.

For generating sequences, the problem is more difficult: the model generates one frame at a time, and the final result must be coherent over long sequences. Recent recurrent neural networks architectures, like \textit{Long-Short Term Memory} (LSTM) \cite{hochreiter1997long} and \textit{Gated Recurrent Units} (GRU) \cite{cho2014learning,chung2014empirical}, achieve unprecedented performance in handling long sequences.

Theses architectures has been used in many applications, like learning language models \cite{sutskever2011generating,Sutskever:2014:SSL:2969033.2969173}, image captioning \cite{karpathy2015deep,vinyals2015show}, music generation \cite{briot2017music} and speech synthesis \cite{oord2016wavenet}.

\par Focus was dedicated to use these powerful tool in order to extract meaningful latent space. One such work that inspired the investigation in this paper is \cite{DBLP:journals/corr/HaE17}. In their work, they investigated the problem of sketch drawing \cite{quickdraw} using a Variational Autoencoder. The latent space emerged encoded meaningful semantic information about these drawings. In our work, we simple a similar architecture, without the variational part, showing that similar behaviour.

\subsection{Data Representation}
\par For handwriting, a continuous coordinate representation (e.g. continuous X, Y) seems the natural option. However, generating continuous data is not straightforward. 
Traditionally, in neural networks, when we want to output a continuous value, a simple linear or \textit{Tanh} activation function is used in the output layer of the neural network. 

However, Bishop~\cite{bishop1994mixture} studied the limitations of these functions and showed that they can not model rich distributions. In particular, when the input can have multiple outputs (one-to-many), these functions will average over all the outputs. He proposed the use of \textit{Gaussian Mixture Model} (GMM) as the final activation function of a neural network. The alliance of neural networks and GMMs is called \textit{Mixture Density Network} (MDN). The training consists in optimizing the GMM parameters (means, covariances). The inference is done by sampling from the GMM distribution. 

To simplify the process, and focus our study on investigating of styles, we extract two features for the tracings: directions and speed (explained in section \ref{sec:preprocessing}), and we quantize these features. Thus, we can model each point in the letter tracings as a categorical distribution, and use a simple \textit{SoftMax} function as the output of the network, which is much simpler than MDN. This was inspired by the studies done in \cite{oord2016wavenet,VanDenOord:2016:PRN:3045390.3045575}, where they report impressive results on originally continuous data, using suitable quantization policy. A categorical distribution is more flexible and generic than continuous ones.
\subsection{Evaluation metrics}
\par The objective evaluation of a generative model is a challenging task, since there is no consensus for objective evaluation metrics. In many cases, a subjective evaluation is performed to overcome this problem. For handwriting of Chinese letters, \cite{DBLP:journals/corr/abs-1801-08624} proposed two metrics:
\begin{description}
    \item[Content accuracy]: They train an \textit{evaluator} model on the ground truth data, and use it to recognize the letters produced by their generator. This approach however faces important problems: the model is trained with ground truth data, and this results in error in the classification, $E_{eval-ref}$. We call the error of the generator $E_{gen}$. When the evaluator is exposed to the data coming from the generator, a new source/distribution of errors is now coming from the generator, which the evaluator have never been exposed to before, leading to a change in the evaluator error behavior. We call this new error $E_{eval-gen}$. Thus, there are no guarantee that the result of the evaluator is faithful in this case. It is also not possible to deduce $E_{gen}$ from just knowing $E_{eval-ref}$ and $E_{eval-gen}$, since the model performance in this case is unknown. 
    \item[Style discrepancy]: In \cite{DBLP:journals/corr/GatysEB15a}, the authors performed image style transfer: take an image, and transform it to the style of an artist. In order to evaluate the quality of the transfer, they measured the correlation between different filter activations (in convolutional neural network) at one layer -- which represent the style representation --. While this metric is interesting to explore, it is not directly applicable to our case, since it assumes the use of convolutional neural network.
\end{description}

\par \cite{mohammed2018DTL} also addressed the problem of evaluation of handwriting generation. They used the \textit{BLEU score} \cite{papineni2002bleu} (a metric widely used in text translation and image captioning) and the \textit{End of Sequence} (EoS) analysis. They showed that these metrics correlate with the quality of the generated letter.

\par The BLEU score is global: all frames of the generated sequence contribute to the final score. The BLEU score is used to compare segments of generated traces with the ground truth. Depending on the number of \textit{grams} chosen, the BLEU score can compare larger segments, thus giving us different levels of granularity to assess the quality of the generated samples.


\par The EoS is a simple yet important style feature. Some letters take longer (e.g., written using many strokes, like H or E) to write than other letters (e.g., written with one stroke like O or C). It is also an idiosyncratic feature of the writer: writers have different writing speeds, depending on age, education or cognitive/peripheral disorders.

\section{Dataset and Pre-Processing}
\subsection{Dataset} \label{sec:data}
\par In this study, we use the \textit{IRON-OFF} Cursive Handwriting Dataset \cite{791823}. This dataset provides us with isolated letters, thus allowing us to focus on the problem of styles with a limited number of strokes per item, unlike other handwriting datasets such as \textit{IAM Handwriting Database} \cite{marti1999full}. To summarize this dataset:

\begin{itemize}
    \item Around 700 writers in total. We use the 412 writers who have written isolated letters.
    \item 10,685 isolated lower case letters, 10,679 isolated upper case letters, 4,086 isolated digits and 410 euro signs.
    \item The gender, handiness (left or right handed), age and nationality of the writers.
    \item For each example (letter, digit, euro sign), we have that example's image - with size around ~167x214 pixels, and a resolution of 300 dpi -, pen movement timed sequence comprising continuous X, Y and pen pressure, and also discrete pen state. This data is sampled at 100 points per seconds on a Wacom UltraPad A4. 
\end{itemize}

\par We focused on the uppercase letters only, and we did not use the pen state or the pen pressure. The idea was to limit number the possible style factors, so that we can better study them.

\par One challenging issue with this dataset however is that we have only one example for each writer-letter combination. This makes the task more difficult, because it is hard to extract a writer style using very few items (the 26 letters/writer in this case).

\subsection{Pre-processing} \label{sec:preprocessing}


\par The letters tracings has been cleaned by removing points related to false starts or corrections as well extra strokes. Tracings with length exceeding 1 second has been removed, as well as tracings more than 99 time steps. This is because they are quite rare, thus, their existence would significantly degrade the performance of our model.

\par We represent each letter tracing by two features: directions and speed. Each feature is quantized into 16 levels and represented as a one-hot encoded vector.

\par Freeman codes~\cite{freeman1961encoding} is used in order to encode the direction feature. It belongs to a family of compression algorithms called \textit{Chain Codes}. This set of algorithms proved to be useful to encode an image with connected components. They can transform a sparse matrix to just a small fraction of the size of the image, in the form of a sequence of codes. Thus, they are being used as compression algorithms as well.

\par Freeman codes can N-directional codes (where N are the directions), depending on the needed resolution. It is quite simple as it encodes each direction with a unique number from 0 to N-1. A direction is defined as the directed vector connecting two neighbouring pixels on the contour of a connected component in the image.

\par
\begin{figure}[htbp!]
\centering
\includegraphics[scale=0.4]{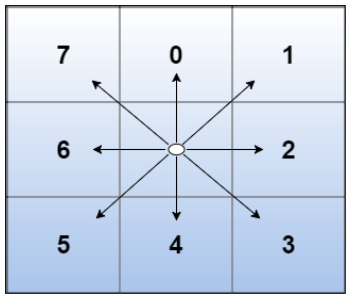}
\caption{Example for freeman code representation for 8 directions. Each direction is given a unique number.}
\label{fig:freeman_dir}
\end{figure}

 We compute the change of directions between three consecutive points. Then, we map this change to its corresponding freeman code number, as shown in figure~\ref{fig:freeman_dir}. Last, we transform the direction number into one-hot encoding scheme, and use this as input to our network. We also quantize the speed of each displacement.

\section{Model architecture}
 \par
\begin{figure}[htbp!]
\centering
\includegraphics[scale=0.7]{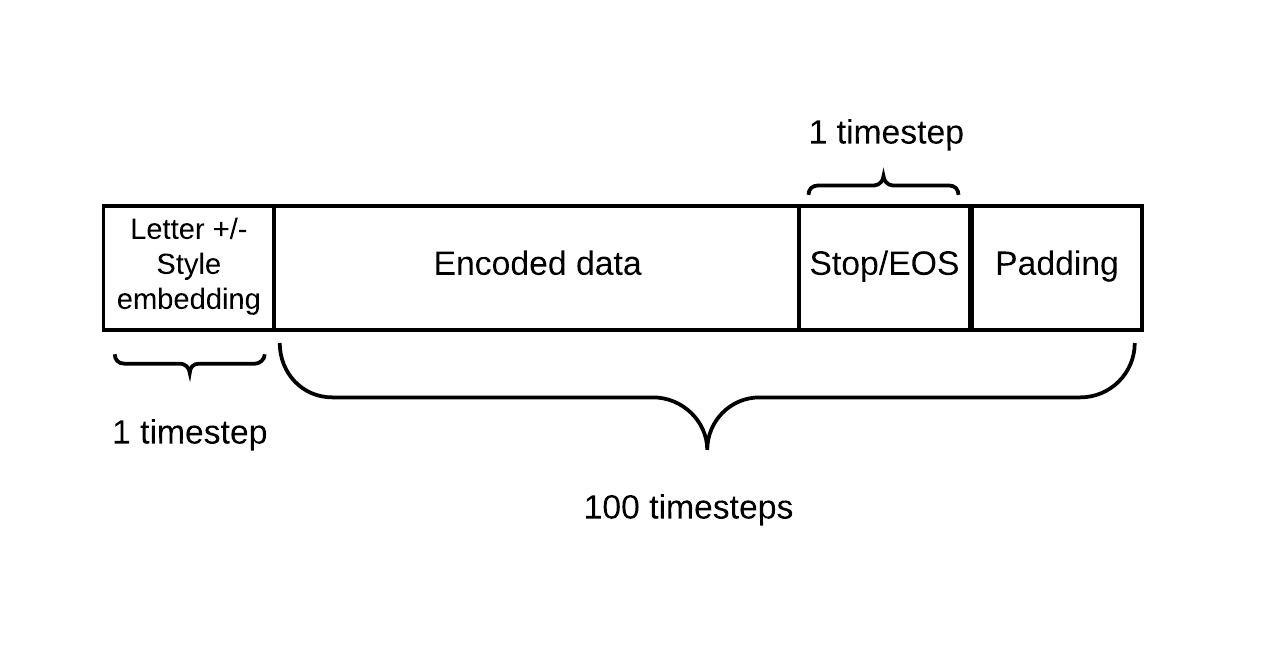}
\caption{Input sequence to our model. The first time step contains the information necessary to condition/bias our model. In case of the encoder, this first time step (the bias) is not included.}
\label{fig:input_shape}
\end{figure}

\begin{figure*}[htbp!]
    \centering
    \begin{subfigure}{1.0\textwidth}
        \centering
        \includegraphics[scale=0.6]{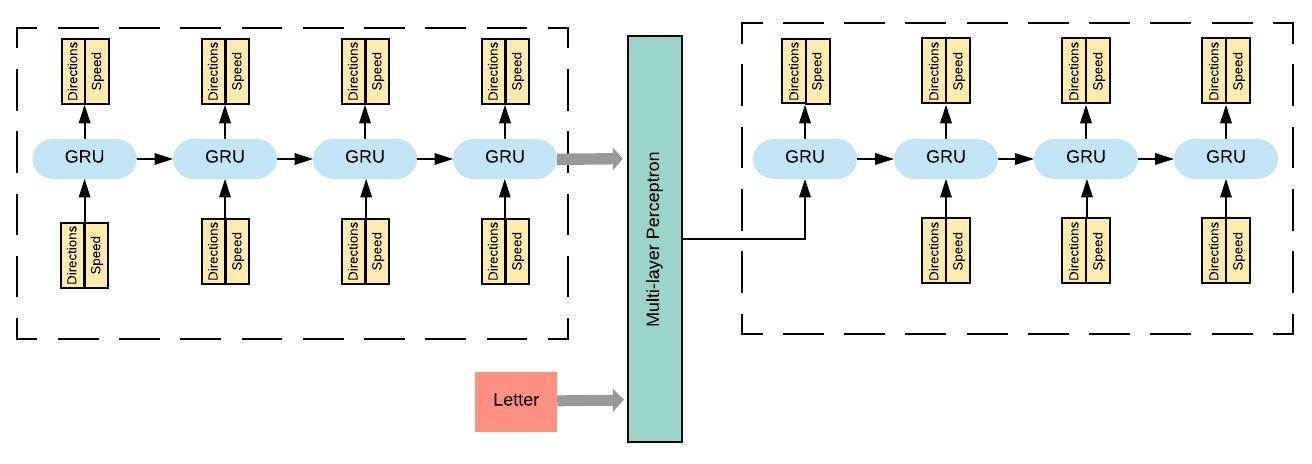}
        \caption{Training mode\label{fig:training_mode}}
    \end{subfigure}
    \begin{subfigure}{1.0\textwidth}
        \centering
        \includegraphics[scale=0.6]{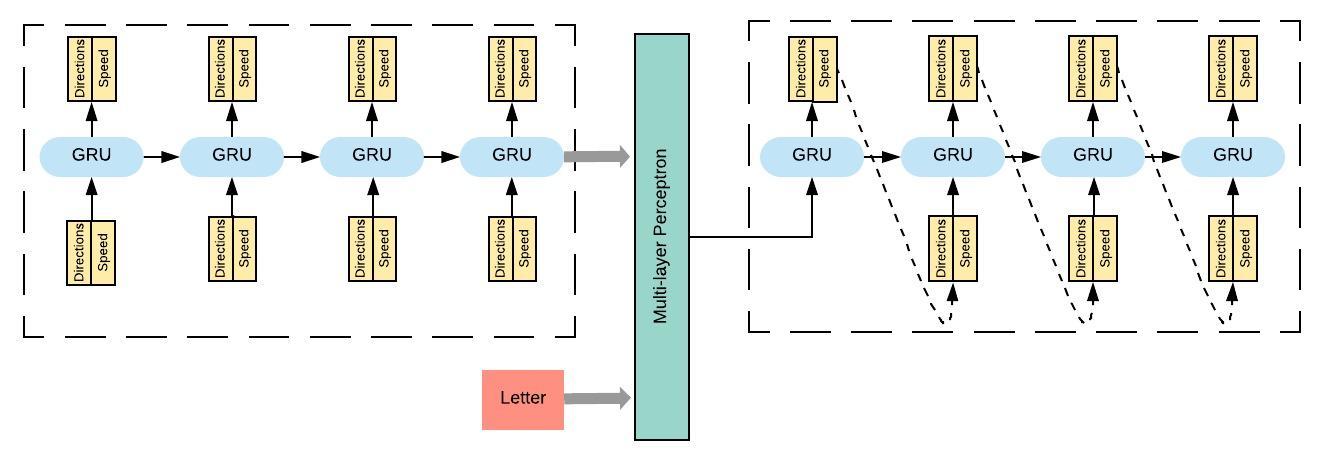}
        \caption{Inference mode\label{fig:inf_mode}}
    \end{subfigure}
    \caption{Schematic diagram of the model we used. During the training time \ref{fig:training_mode}, the input to the model is always the ground truth. During the inference time \ref{fig:inf_mode} however, the input to the decoder (generator) part at each time step is its own predication in the previous time step.}
    \label{fig:model_arch}
\end{figure*}

 The model architecture is illustrated in figure~\ref{fig:model_arch}. The input/output frames of the model are detailed in figure~\ref{fig:input_shape}. The trace of the letter is first fed to encoder module. The final hidden state of that module summarizes the letter. In order to allow this module to focus on learning the style embedding, we complement this last hidden state with the one-hot encoding of the letter identity, and use a projection of them as the bias input to the generator. Thus, we decouple the \textit{task space} -- the letter -- from the \textit{style space}: the encoder is free from the need to learn the letter identity, and can focus learning additional information that enables the generator to better approximate the ground truth tracings.
\par In the decoder, we follow the framework proposed by~\cite{vinyals2015show} in order to bias the model: we create an extra time step at the beginning, which has the information we want to bias the model with. In this case, this time step is the projection of the encoder last hidden state and the letter encoder. This has a much lower dimension than encoder hidden state (the hyperparameters are discussed in section \ref{hyperparam}). This further encourage the model to learn only necessary style information, as suggested in \cite{DBLP:journals/corr/abs-1803-09047}.

\subsection{Hyper-parameter tuning} \label{hyperparam}
\par We ran random hyper-parameter search for a wide range of parameters (learning rate, size and the number of layers for the encoder and the decoder, dropout percentage, etc). GRU layers \cite{cho2014learning,chung2014empirical} is being used in this model. We use \textit{Adam} \cite{kingma2014adam} optimizer in this work, with a fixed learning rate. The implementation is done using \textit{PyTorch} framework \cite{paszke2017automatic}.

\par In order to allow for faster exploration of different hyper-parameters, we use an early stopping of 20 epochs (no improvement happens during these epochs). To summarize, the current model specifications:
\begin{itemize}
    \item Encoder hidden size: 128
    \item Decoder hidden size: 128
    \item Encoder layers: 2
    \item Decoder layers: 2
    \item Encoder dropout: 0.0
    \item Decoder dropout: 0.2
    \item Learning rate: 0.001
\end{itemize}

\subsection{Training}
\par The encoder and the decoder parts have the target of modeling the next time step in the sequence, $x_{t+1}$, given the previous time steps, or in other words, $P(x_{t+1}|x_1, x_2,...,x_{T})$, where $x_t$ is the tracing point at time $t$, and $T$ is the length of the input sequence (see figure \ref{fig:input_shape}).  To achieve this, the model is given the ground truth input of points $x1, x2,..., x_{T-1}$ and is asked to output the sequence $x2, x3,..., x_{T}$.

\par The model is trained to minimize the negative log likelihood loss of the correct point at each time step. For each feature (speed and freeman codes), it is calculated as in equation \ref{eq:loss_one_modality}. The final loss is the average loss of the two feature, as in equation \ref{eq:total_loss}.
\begin{equation}
    \begin{split}
    Loss = - \log \prod_{t=1}^{T} p(x_{t}|x_1, x_2,...,x_{t-1}) \\
    = - \sum_{t=1}^{T} \log p(x_{t}|x_1, x_2,...,x_{t-1})
    \label{eq:loss_one_modality}
    \end{split}
\end{equation}
\begin{equation}
    Total Loss = (Loss_{speed} + Loss_{freeman}) / 2.0
    \label{eq:total_loss}
\end{equation}

\par During the training, the output of the model at each time step is the:
\begin{equation}
    x^{g}_{t+1} = {argmax}_{x}p(x|x_{t},h_{t})
\end{equation}
where $x^{g}_{t+1}$ is the generated/predicted next time step by the model, $x_{t}$ is the ground truth input at the current time step $t$, and $h_{t}$ is the hidden state of the GRU at the current time step. Thus, during the training, the model is exposed only to the ground truth data as input.

\subsection{Inference}

\par To sample from the model, we used the \textit{softmax sampling} strategy: fit the output of the network into two multinomial distributions (one for freeman codes and the other for the speed). We then sample the next time step from these two distributions. We can control the level of randomness of the sampling using a \textit{temperature} parameter for the softmax function. We tried different temperatures, and we found the value of $0.5$ achieves the best results. The generation continues till $N_{max}$ time steps - which is 100 time steps in our case -.

\section{Evaluation Metrics} \label{sec:evaluation}
\par Evaluation is a challenging problem when using generative models. We want metrics to capture the distance between the generated and the ground truth distributions. Following the work done in \cite{mohammed2018DTL}, we use the same two evaluation metrics in our model:
\begin{itemize}
    \item \textbf{BLEU score}~\cite{papineni2002bleu} It is a well known metric to evaluate text generation applications, like image captioning \cite{karpathy2015deep,vinyals2015show} and machine translation \cite{Sutskever:2014:SSL:2969033.2969173}. Since we discretized the letter drawings, this fits nicely within our work. The general intuition is the following: if we take a segment from the generated letter, did this segment happen in the ground truth letter? We keep doing this for segments of increasing length (the length of the segment here is the number of grams used in the BLEU score). For our work, we report the results on segments from 1 to 3 time steps.
    
    Each part of the letter has two parallel segments: freeman codes and speed, thus, we report the BLEU score for both of them. 
    The equation to compute the BLEU score is the following:
    \begin{equation}
    BLEU_{N} = \frac{\sum_{C\in G}\sum_{N\in C}Count_{Clipped}(N)}{\sum_{C\in G}\sum_{N\in C}Count(N)}
    \end{equation}
    \begin{equation}
    Score_{N} = \min{(0, 1 - \frac{L_{R}}{L_{G}})} \prod^{N}_{n=1}BLEU_{n}
    \end{equation}
    
    where: $G$ is all the generated sequences, $N$ is the total number of N-grams we want to consider. $Count_{Clipped}$ is clipped N-grams count (if the number of N-grams in the generate sequence is larger than the reference sequence, the count is limited to the number in the reference sequence only), $L_R$ is the length of the reference sequence, $L_G$ is the length of the generated sequence. 
    The term $\min(0, 1 - \frac{L_{R}}{L_{G}})$ is added in order to penalize short generated sequences (shorter than the reference sequence), which will deceptively achieve high scores.
    
    \item \textbf{End of Sequence} The length of the letter is another aspect of the style. The distribution of length in the generated examples should follow the ground truth examples. In order to perform this analysis, we compute ~\textit{Pearson correlation coefficient} between the generated examples and the ground truth data.
\end{itemize}

\section{Experiments and results}

\subsection{Letter generation with style preservation}
\par The objective here to compare the quality of the generated letters to the state-of-the-art benchmarks. As mentioned earlier, we compare using the BLEU score metric and the EoS analysis. 
The BLEU score results can be seen in table \ref{table:bleu_gen}, and the results for EoS analysis results are in table \ref{table:EoS_gen}. We can see that the BLEU-3 score results of our model achieves 32.3\% accuracy in Speed feature and 38.7\% accuracy in Freeman feature, compared to 25.1\% and 28.3\% accuracy using the benchmark model on both features respectively.
\par The same goes for the EoS analysis. In comparing the Person Coefficient, our model achieves 0.99 score compared to 0.55 for the benchmark model (the highest score is 1.0). This is a support that our model capture the style of handwriting better than the benchmark. 
\par
\begin{figure*}[!htbp]
\centering
    \begin{subfigure}[b]{0.15\textwidth}
        \includegraphics[width=\textwidth]{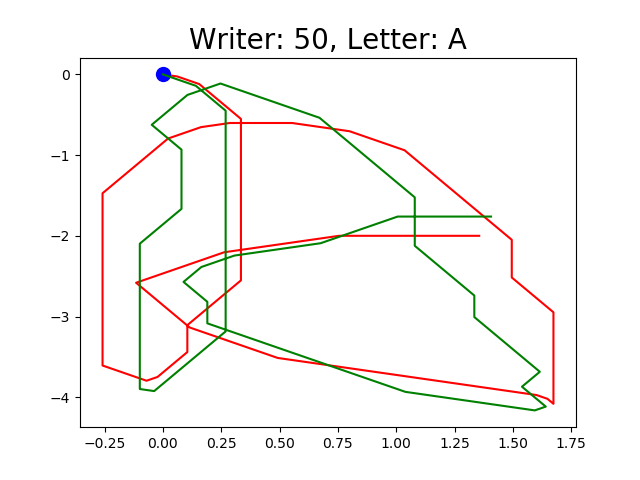}
    \end{subfigure}
    ~ 
    \begin{subfigure}[b]{0.15\textwidth}
        \includegraphics[width=\textwidth]{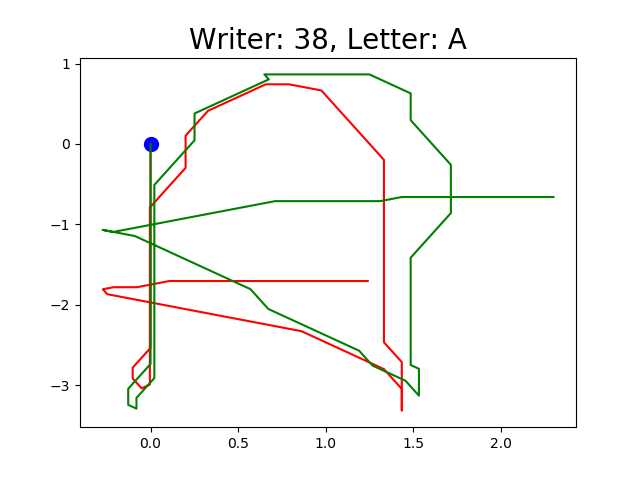}
    \end{subfigure}
    ~
    \begin{subfigure}[b]{0.15\textwidth}
        \includegraphics[width=\textwidth]{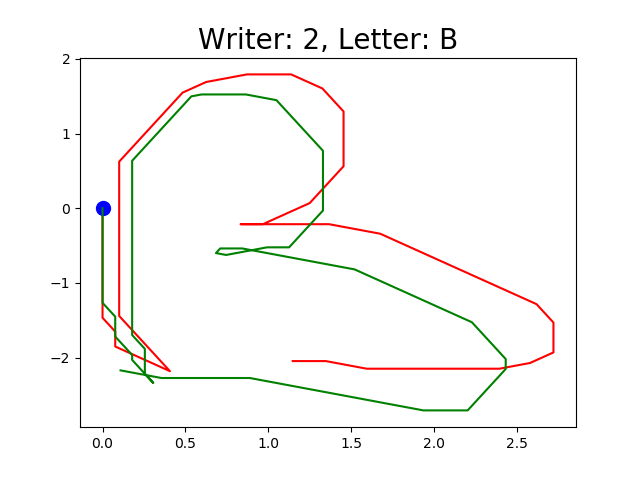}
    \end{subfigure}
    ~
    \begin{subfigure}[b]{0.15\textwidth}
        \includegraphics[width=\textwidth]{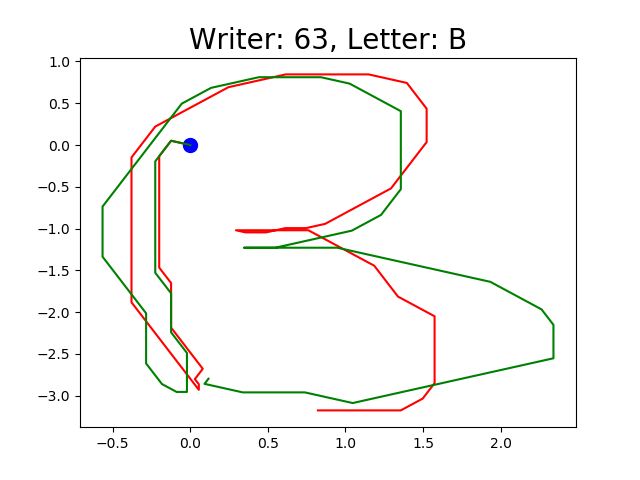}
    \end{subfigure}
    ~
    \begin{subfigure}[b]{0.15\textwidth}
        \includegraphics[width=\textwidth]{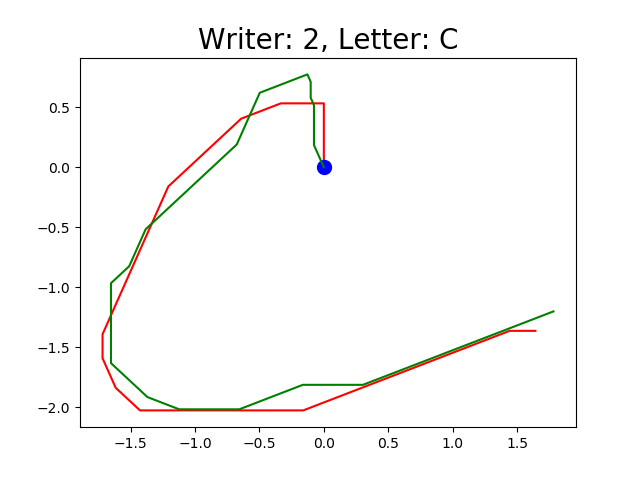}
    \end{subfigure}
    ~
    \begin{subfigure}[b]{0.15\textwidth}
        \includegraphics[width=\textwidth]{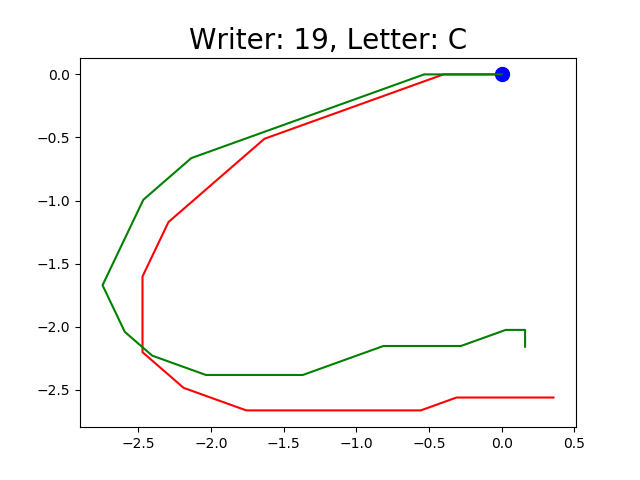}
    \end{subfigure}
    ~
    \begin{subfigure}[b]{0.15\textwidth}
        \includegraphics[width=\textwidth]{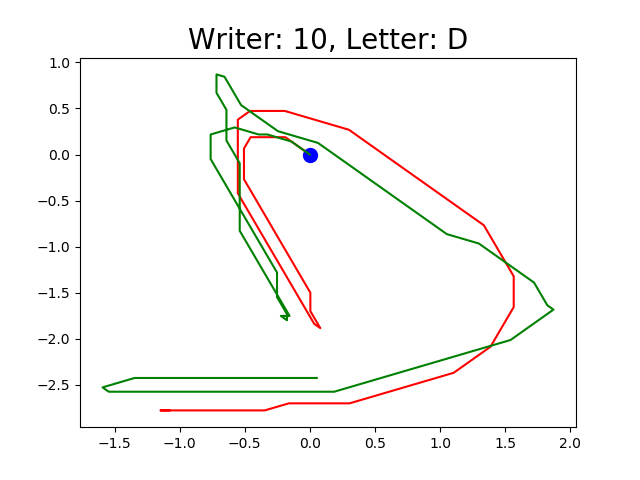}
    \end{subfigure}
    ~
    \begin{subfigure}[b]{0.15\textwidth}
        \includegraphics[width=\textwidth]{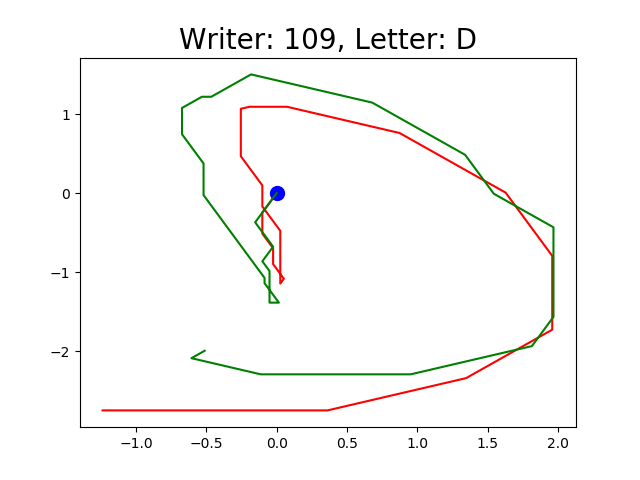}
    \end{subfigure}
    ~
    \begin{subfigure}[b]{0.15\textwidth}
        \includegraphics[width=\textwidth]{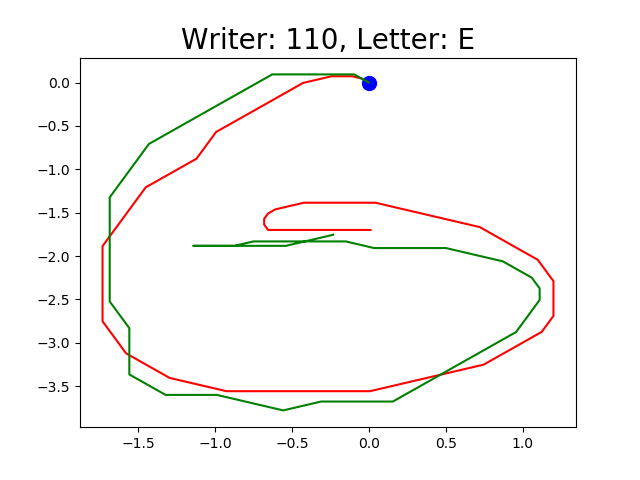}
    \end{subfigure}
    ~
    \begin{subfigure}[b]{0.15\textwidth}
        \includegraphics[width=\textwidth]{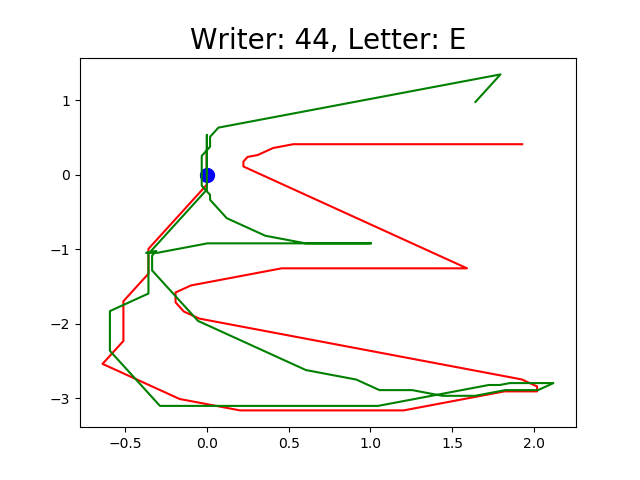}
    \end{subfigure}
    ~
    \begin{subfigure}[b]{0.15\textwidth}
        \includegraphics[width=\textwidth]{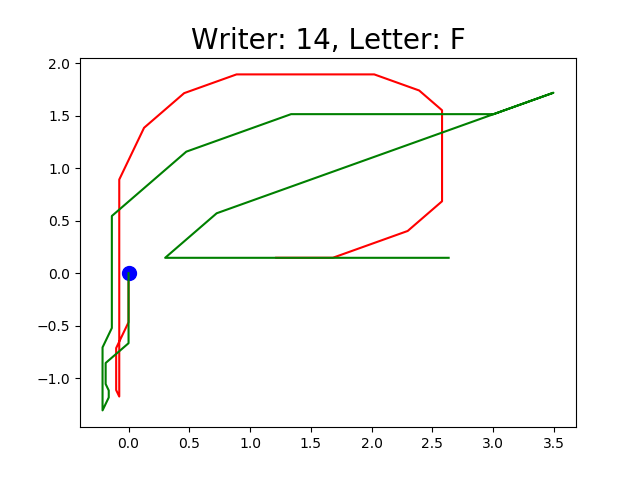}
    \end{subfigure}
    ~
    \begin{subfigure}[b]{0.15\textwidth}
        \includegraphics[width=\textwidth]{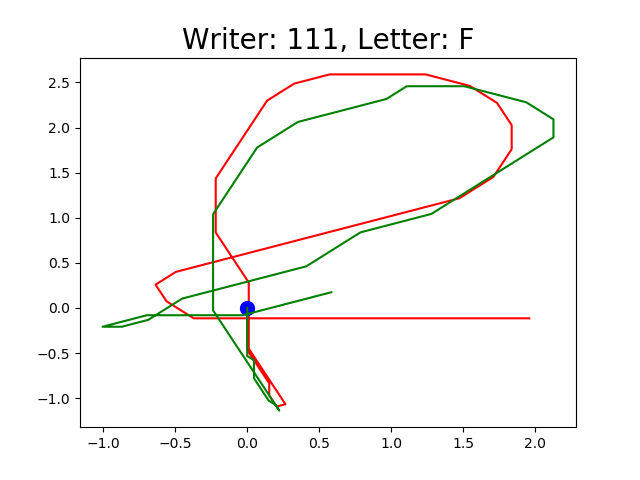}
    \end{subfigure}
    ~
    \begin{subfigure}[b]{0.15\textwidth}
        \includegraphics[width=\textwidth]{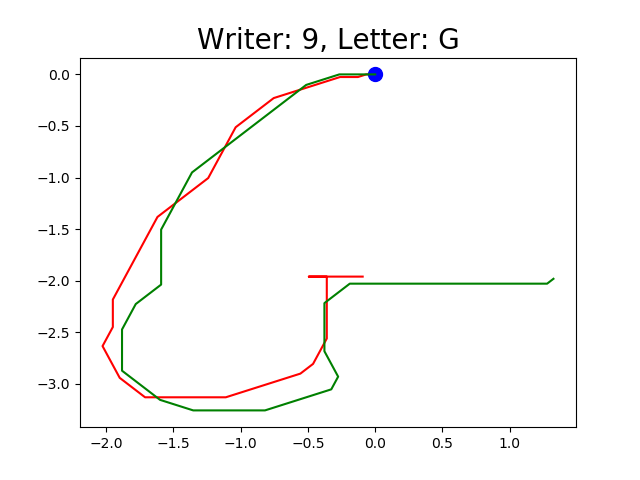}
    \end{subfigure}
    ~
    \begin{subfigure}[b]{0.15\textwidth}
        \includegraphics[width=\textwidth]{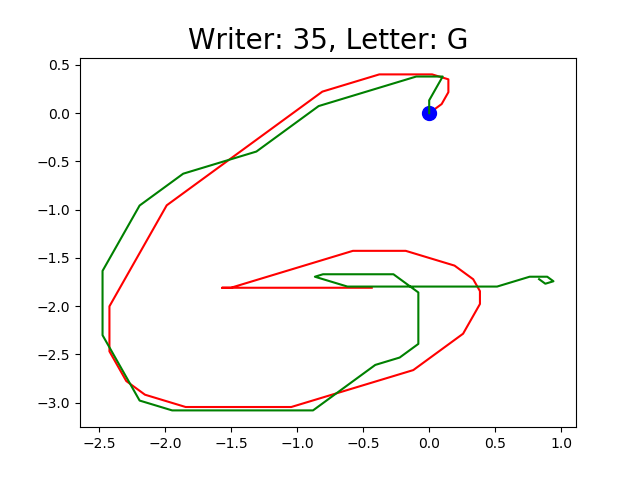}
    \end{subfigure}
    ~
    \begin{subfigure}[b]{0.15\textwidth}
        \includegraphics[width=\textwidth]{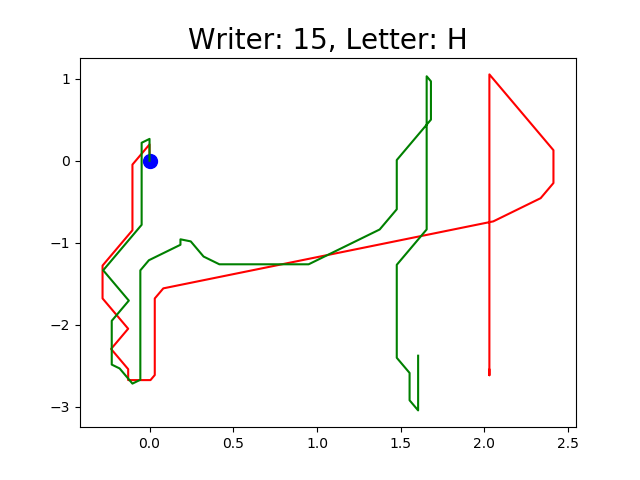}
    \end{subfigure}
    ~
    \begin{subfigure}[b]{0.15\textwidth}
        \includegraphics[width=\textwidth]{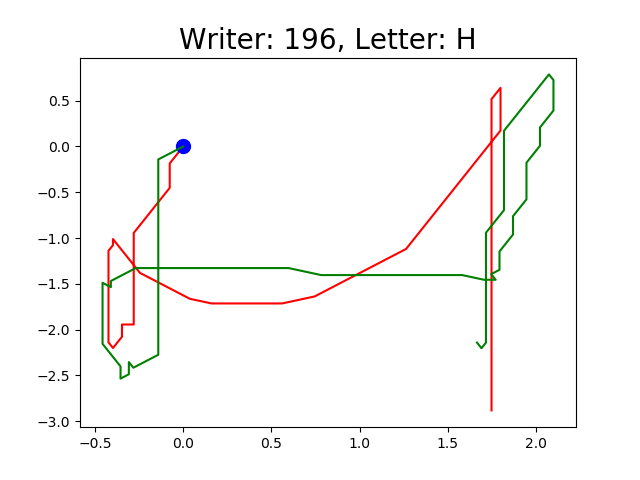}
    \end{subfigure}
    ~
    \begin{subfigure}[b]{0.15\textwidth}
        \includegraphics[width=\textwidth]{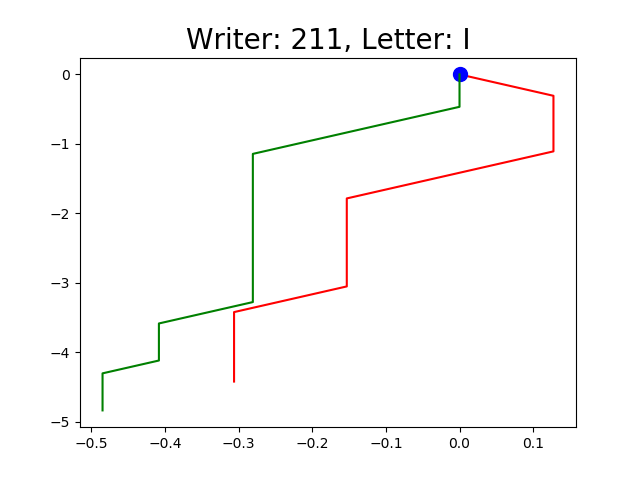}
    \end{subfigure}
    ~
    \begin{subfigure}[b]{0.15\textwidth}
        \includegraphics[width=\textwidth]{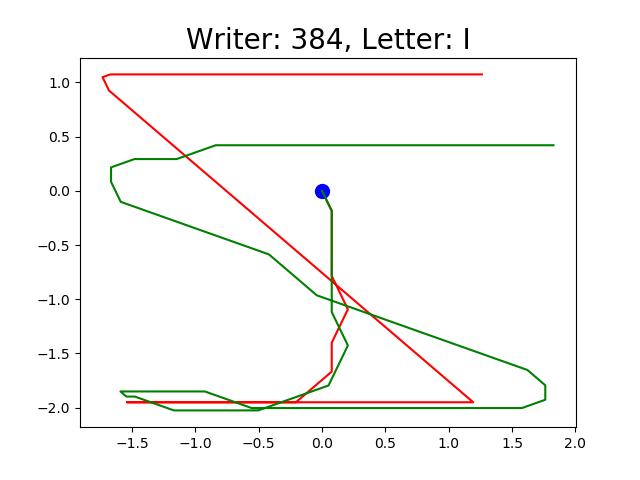}
    \end{subfigure}
    ~
    \begin{subfigure}[b]{0.15\textwidth}
        \includegraphics[width=\textwidth]{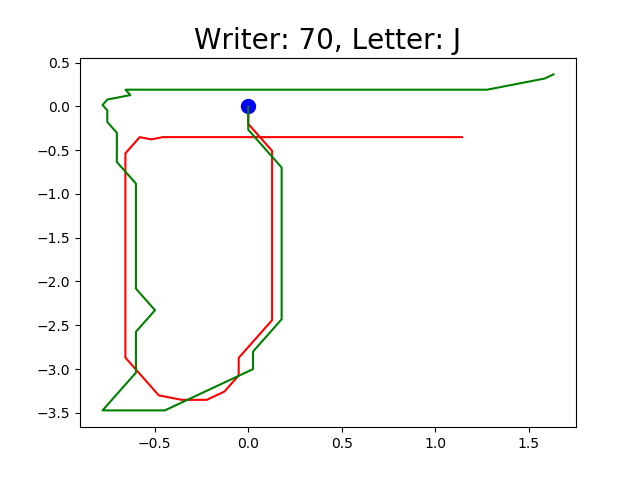}
    \end{subfigure}
    ~
    \begin{subfigure}[b]{0.15\textwidth}
        \includegraphics[width=\textwidth]{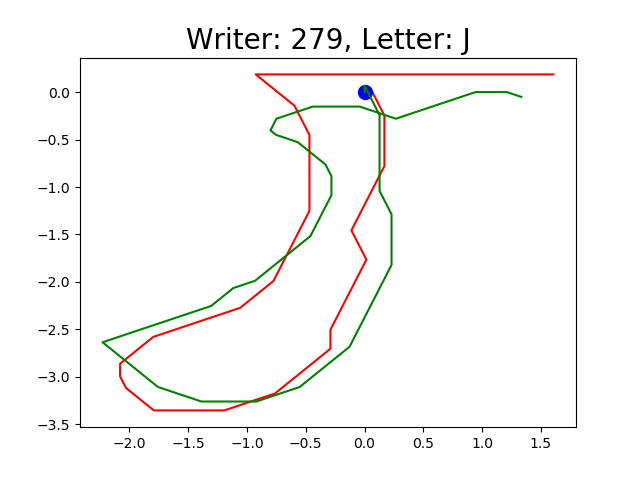}
    \end{subfigure}
    ~
    \begin{subfigure}[b]{0.15\textwidth}
        \includegraphics[width=\textwidth]{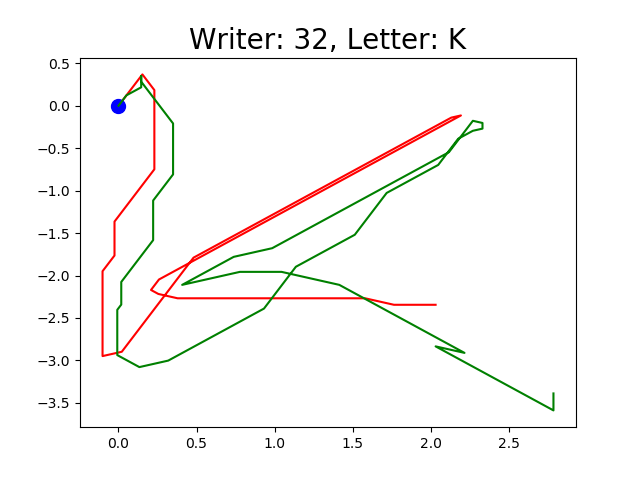}
    \end{subfigure}
    ~
    \begin{subfigure}[b]{0.15\textwidth}
        \includegraphics[width=\textwidth]{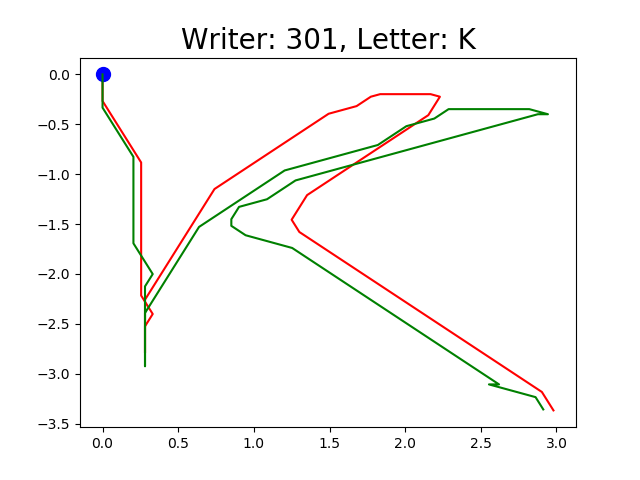}
    \end{subfigure}
    ~
    \begin{subfigure}[b]{0.15\textwidth}
        \includegraphics[width=\textwidth]{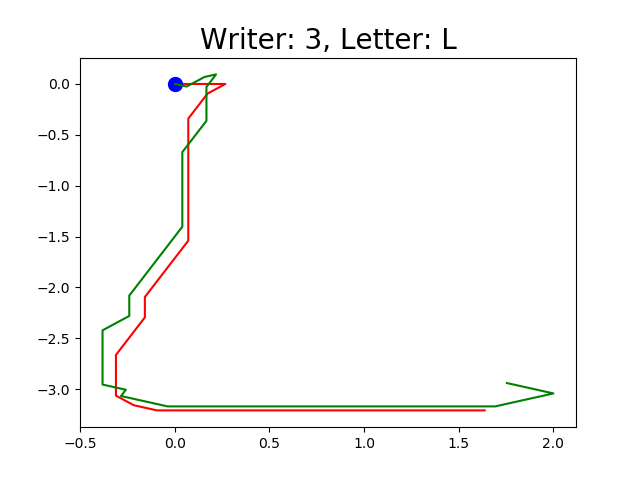}
    \end{subfigure}
    ~
    \begin{subfigure}[b]{0.15\textwidth}
        \includegraphics[width=\textwidth]{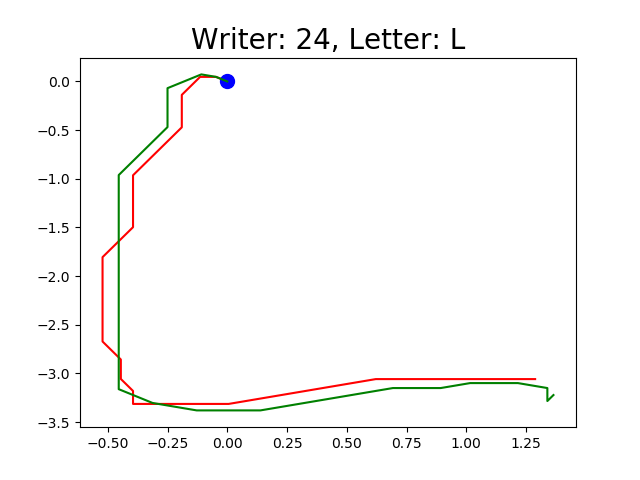}
    \end{subfigure}
    ~
    \begin{subfigure}[b]{0.15\textwidth}
        \includegraphics[width=\textwidth]{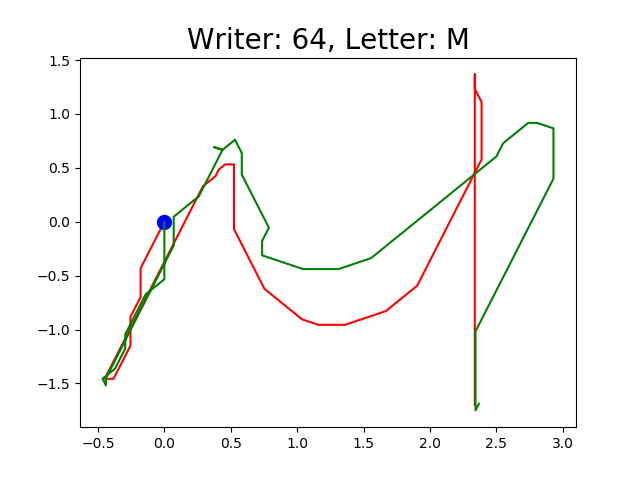}
    \end{subfigure}
    ~
    \begin{subfigure}[b]{0.15\textwidth}
        \includegraphics[width=\textwidth]{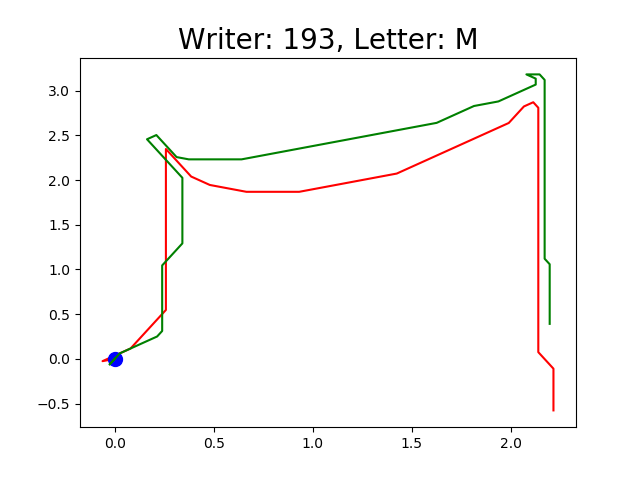}
    \end{subfigure}
    ~
    \begin{subfigure}[b]{0.15\textwidth}
        \includegraphics[width=\textwidth]{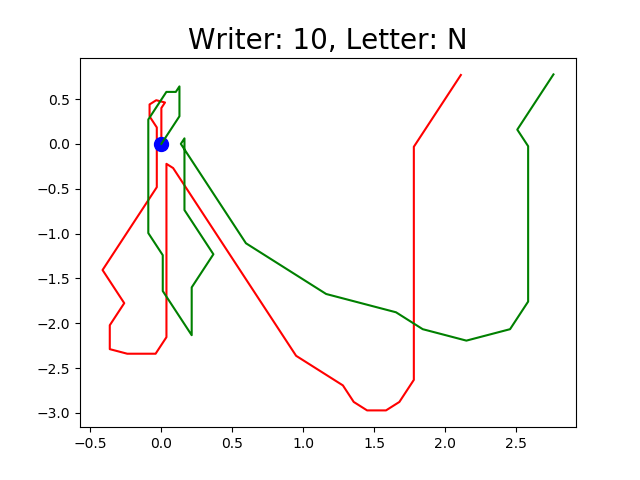}
    \end{subfigure}
    ~
    \begin{subfigure}[b]{0.15\textwidth}
        \includegraphics[width=\textwidth]{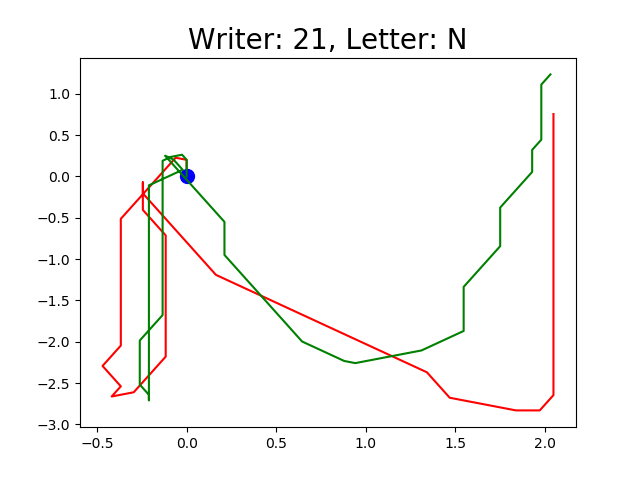}
    \end{subfigure}
    ~
    \begin{subfigure}[b]{0.15\textwidth}
        \includegraphics[width=\textwidth]{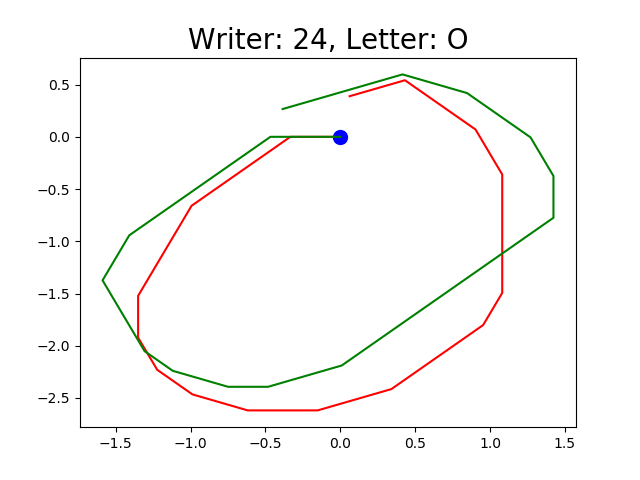}
    \end{subfigure}
    ~
    \begin{subfigure}[b]{0.15\textwidth}
        \includegraphics[width=\textwidth]{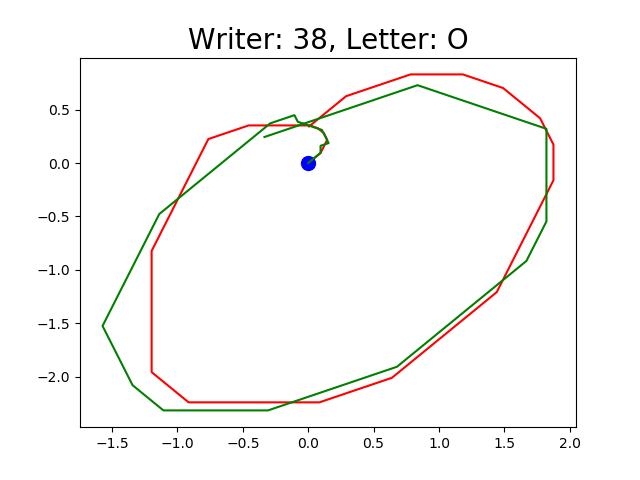}
    \end{subfigure}
    ~
    \begin{subfigure}[b]{0.15\textwidth}
        \includegraphics[width=\textwidth]{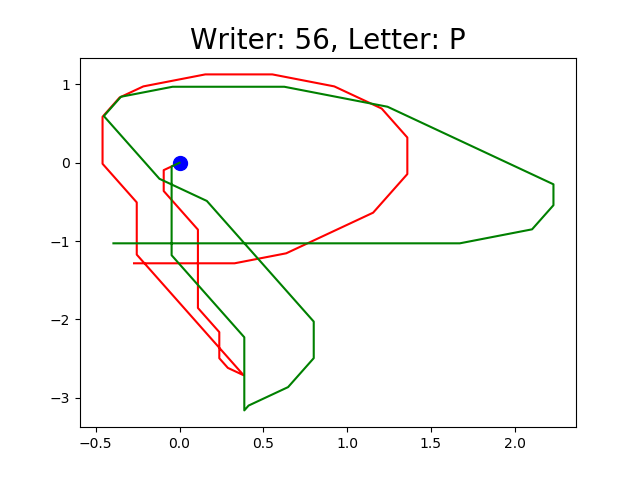}
    \end{subfigure}
    ~
    \begin{subfigure}[b]{0.15\textwidth}
        \includegraphics[width=\textwidth]{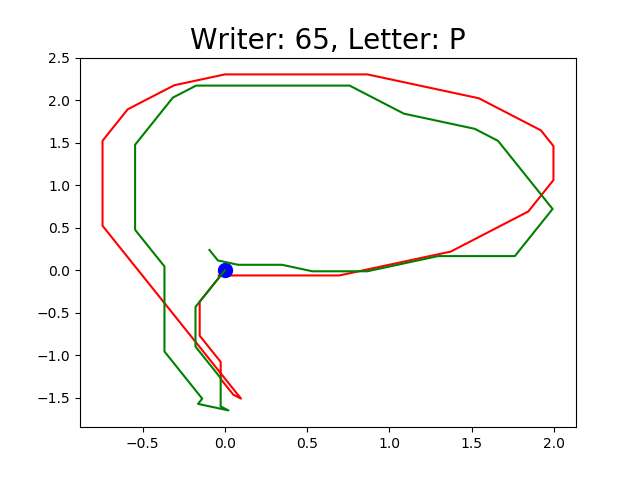}
    \end{subfigure}
    ~
    \begin{subfigure}[b]{0.15\textwidth}
        \includegraphics[width=\textwidth]{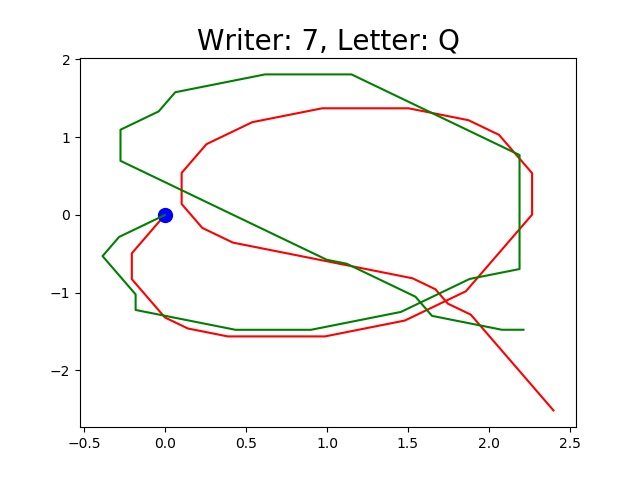}
    \end{subfigure}
    ~
    \begin{subfigure}[b]{0.15\textwidth}
        \includegraphics[width=\textwidth]{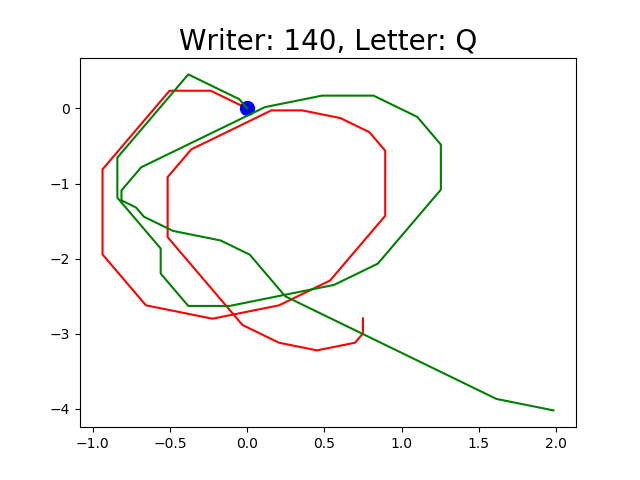}
    \end{subfigure}
    ~
    \begin{subfigure}[b]{0.15\textwidth}
        \includegraphics[width=\textwidth]{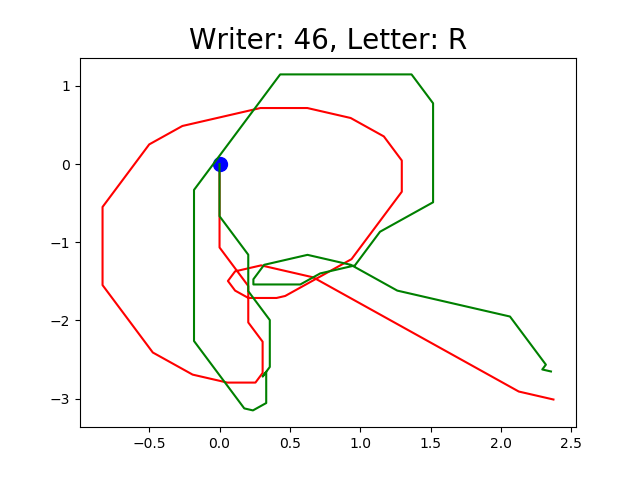}
    \end{subfigure}
    ~
    \begin{subfigure}[b]{0.15\textwidth}
        \includegraphics[width=\textwidth]{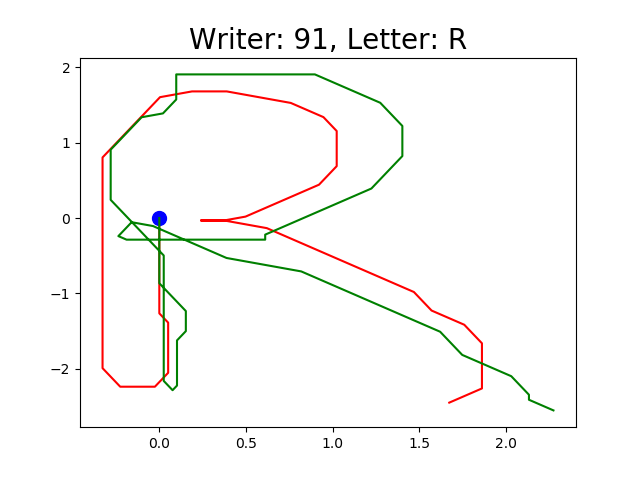}
    \end{subfigure}
    ~
    \begin{subfigure}[b]{0.15\textwidth}
        \includegraphics[width=\textwidth]{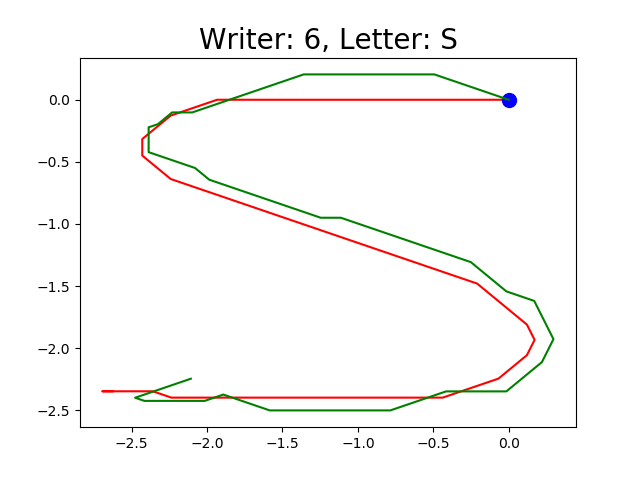}
    \end{subfigure}
    ~
    \begin{subfigure}[b]{0.15\textwidth}
        \includegraphics[width=\textwidth]{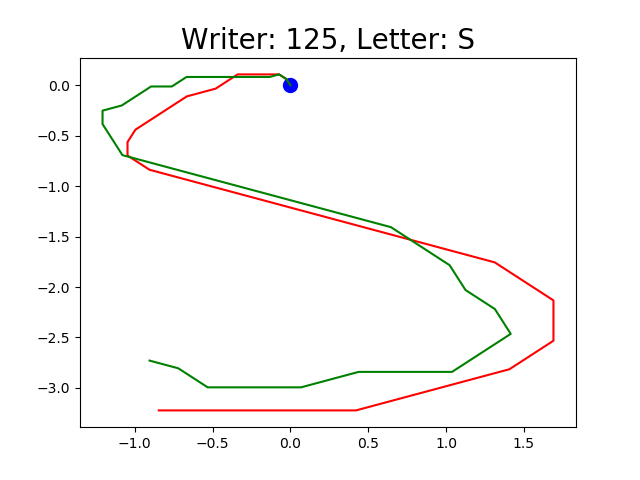}
    \end{subfigure}
    ~
    \begin{subfigure}[b]{0.15\textwidth}
        \includegraphics[width=\textwidth]{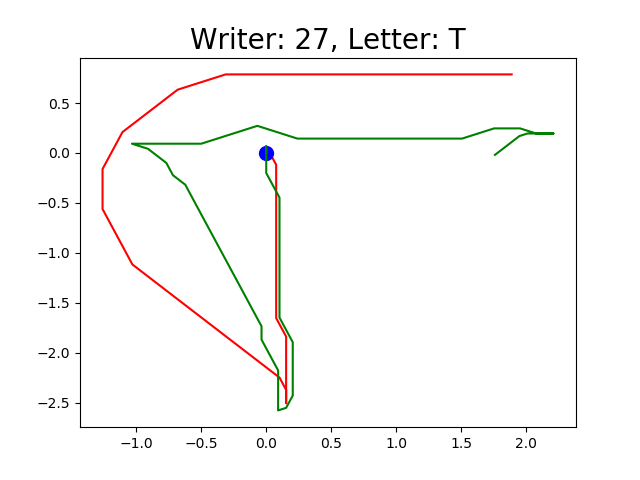}
    \end{subfigure}
    ~
    \begin{subfigure}[b]{0.15\textwidth}
        \includegraphics[width=\textwidth]{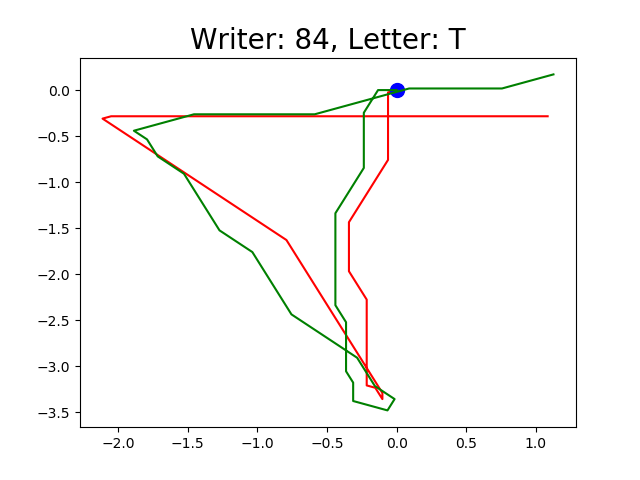}
    \end{subfigure}
    ~
    \begin{subfigure}[b]{0.15\textwidth}
        \includegraphics[width=\textwidth]{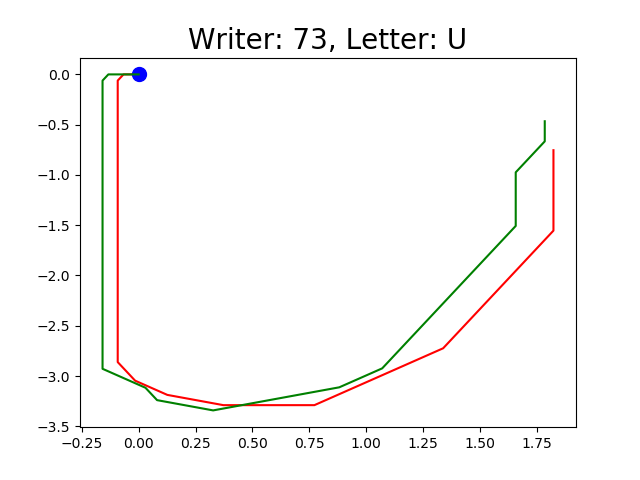}
    \end{subfigure}
    ~
    \begin{subfigure}[b]{0.15\textwidth}
        \includegraphics[width=\textwidth]{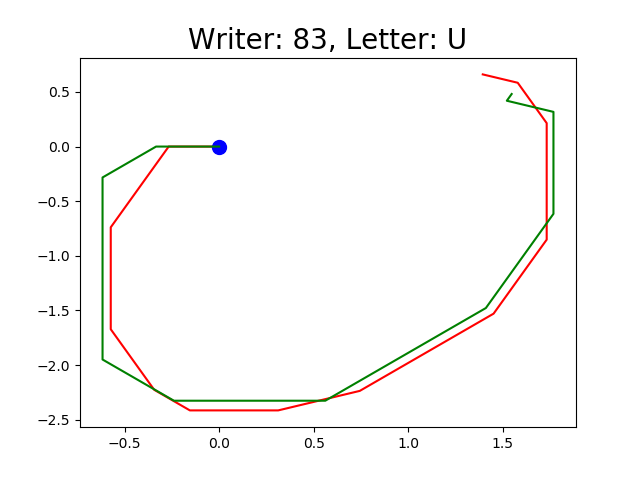}
    \end{subfigure}
    ~
    \begin{subfigure}[b]{0.15\textwidth}
        \includegraphics[width=\textwidth]{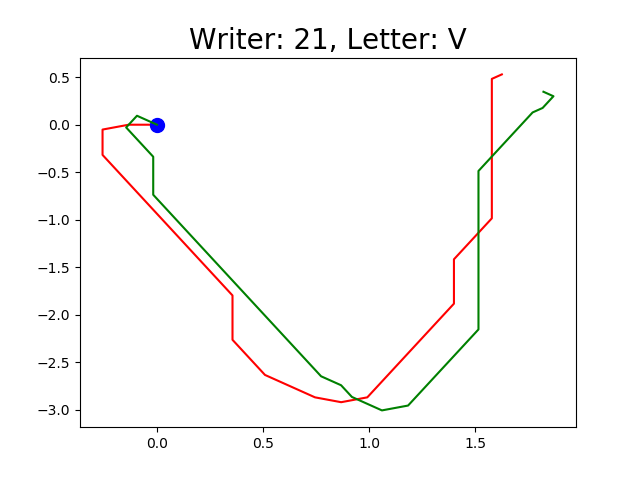}
    \end{subfigure}
    ~
    \begin{subfigure}[b]{0.15\textwidth}
        \includegraphics[width=\textwidth]{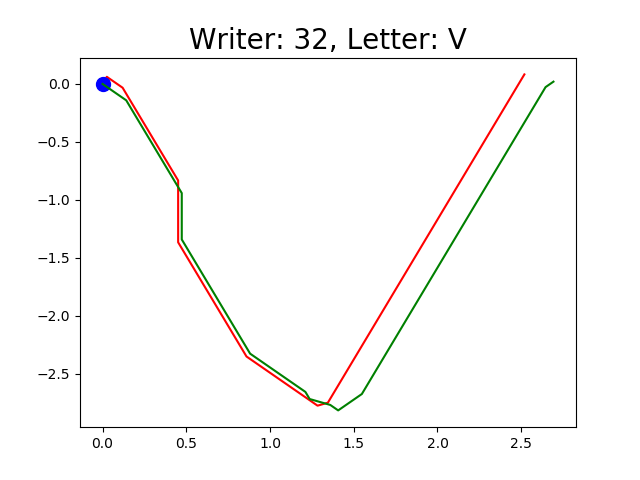}
    \end{subfigure}
    ~
    \begin{subfigure}[b]{0.15\textwidth}
        \includegraphics[width=\textwidth]{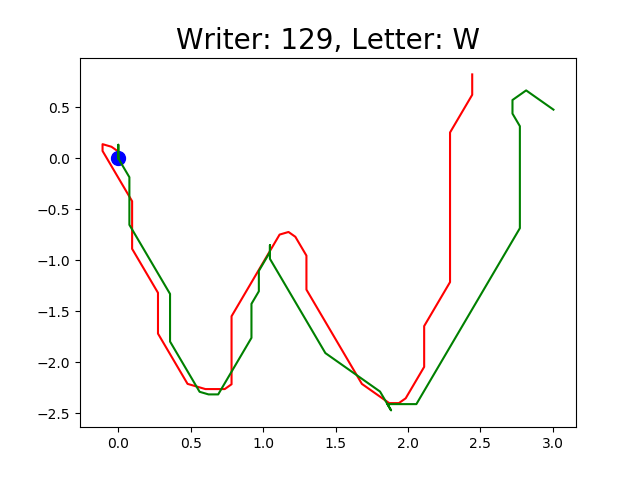}
    \end{subfigure}
    ~
    \begin{subfigure}[b]{0.15\textwidth}
        \includegraphics[width=\textwidth]{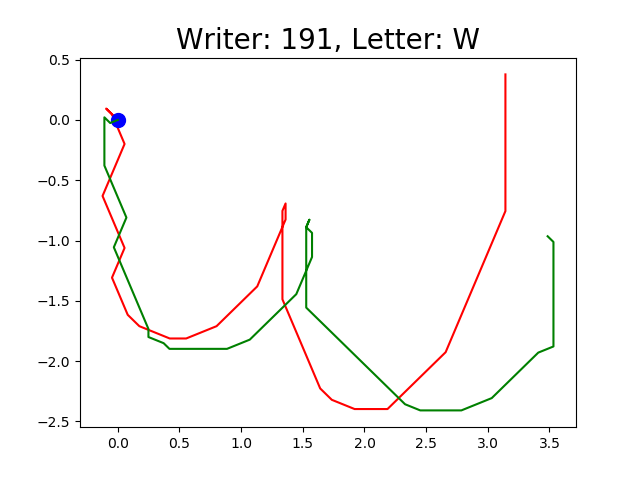}
    \end{subfigure}
    ~
    \begin{subfigure}[b]{0.15\textwidth}
        \includegraphics[width=\textwidth]{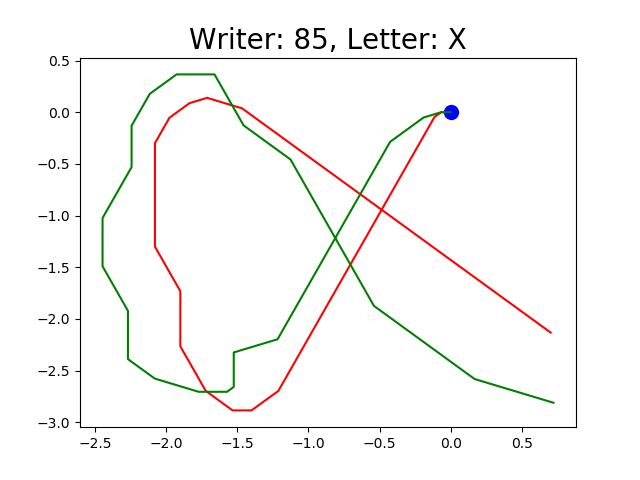}
    \end{subfigure}
    ~
    \begin{subfigure}[b]{0.15\textwidth}
        \includegraphics[width=\textwidth]{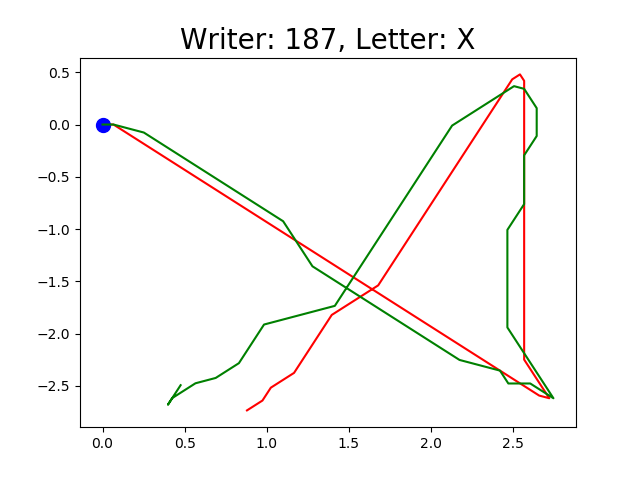}
    \end{subfigure}
    ~
    \begin{subfigure}[b]{0.15\textwidth}
        \includegraphics[width=\textwidth]{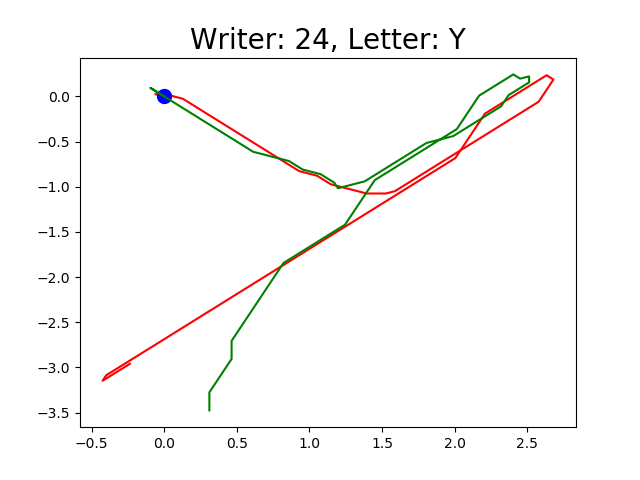}
    \end{subfigure}
    ~
    \begin{subfigure}[b]{0.15\textwidth}
        \includegraphics[width=\textwidth]{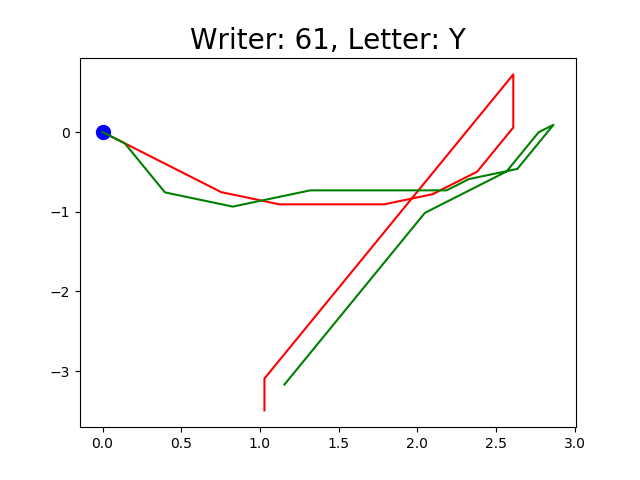}
    \end{subfigure}
    ~
    \begin{subfigure}[b]{0.15\textwidth}
        \includegraphics[width=\textwidth]{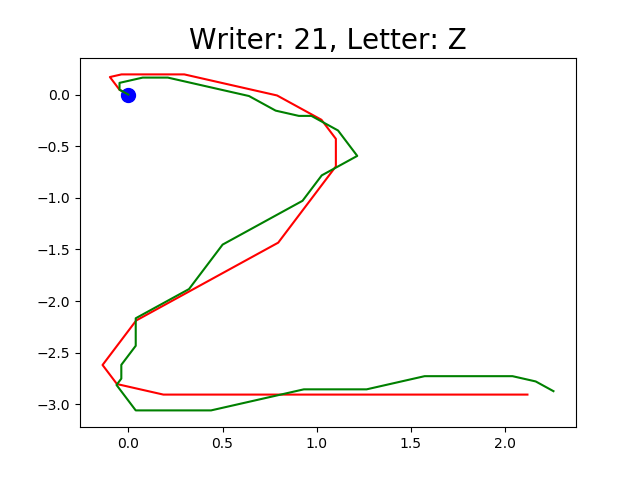}
    \end{subfigure}
    ~
    \begin{subfigure}[b]{0.15\textwidth}
        \includegraphics[width=\textwidth]{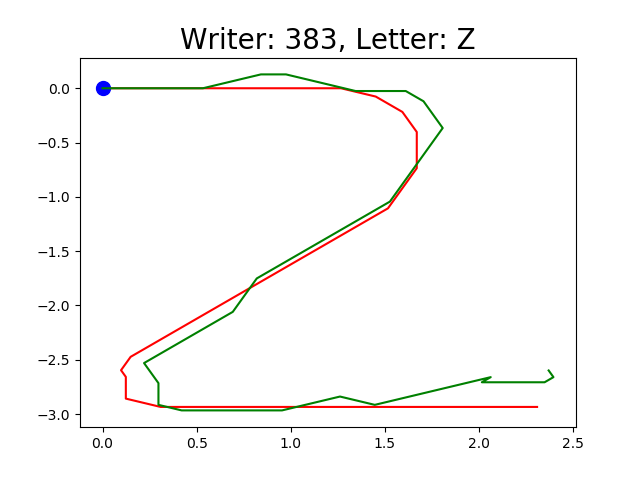}
    \end{subfigure}
    \caption{Examples of generated letters. The blue mark is the starting point. The traces in green is the ground truth, and the red is the generated ones by our model.}
    \label{fig:letters_examples}
\end{figure*}

 Examples for the generated letters can be found in figure \ref{fig:letters_examples}.

\begin{table*}[!htbp]
\centering
\begin{tabular}{|l||c|c|c||c|c|c|} 
\hline
\multicolumn{1}{|c||}{Aspect/Feature} & \multicolumn{3}{c||}{ Speed } & \multicolumn{3}{c|}{ Freeman }   \\ \hline
Model / B-score      & B-1  & B-2  & B-3           & B-1  & B-2   & B-3              \\ \hline
Letter + Writer bias & 51.5 & 41.4 & 25.1          & 56.7 & 39.4  & 28.3             \\\hline
\textbf{Style Extractor} & 71 & 51.7 & 32.3 & 65.6 & 51.5 & 38.7 \\\hline
\end{tabular}
\caption{BLEU scores for different models for known writers.}
\label{table:bleu_gen}
\end{table*}

\begin{table*}[!htbp]
\centering
\begin{tabular}{|l||c|c|c||c|c|c|} 
\hline
\multicolumn{1}{|c||}{Aspect/Feature} & \multicolumn{3}{c||}{ Speed } & \multicolumn{3}{c|}{ Freeman }   \\ \hline
Model / B-score      & B-1  & B-2  & B-3           & B-1  & B-2   & B-3              \\ \hline
Letter + Writer bias & 55.4 & 39.6 & 25.3 & 50.2 & 38.6 & 27.7             \\\hline
\textbf{Style Extractor} & 72.4 & 52.4 & 32.2 & 70.4 & 55.6 & 42.1 \\\hline

\end{tabular}
\caption{BLEU scores for different models for style extraction for 30 new writers (style transfer).}
\label{table:bleu_transfer}
\end{table*}


\begin{table}[!htbp]
\centering
\begin{tabular}{|l|c|c|}
\hline
Models & Pearson coefficient\\ \hline
Letter + Writer bias & 0.55\\ \hline
\textbf{Style Extractor} & 0.99 \\ \hline
\end{tabular}
\caption{Pearson correlation coefficients for the End-Of-Sequence (EoS) distributions for the different models on the normal generation scenario}
\label{table:EoS_gen}
\end{table}


\begin{table}[!htbp]
\centering
\begin{tabular}{|l|c|c|}
\hline
Models & Pearson coefficient\\ \hline
Letter + Writer bias & 0.5\\ \hline
\textbf{Style Extractor} & 0.99\\ \hline
\end{tabular}
\caption{Pearson correlation coefficients for the End-Of-Sequence (EoS) distributions for the different models on 30 new writers (style transfer).}
\label{table:EoS_transfer}
\end{table}

\subsection{Style transfer}
\par One of the hypotheses we want to test is whether there is a limited number of styles needed, to generalize over new writers. To achieve this, the learned representation for styles should extract generic information about the styles. 

\par In order to test this hypothesis, we expose our model to 30 writers that have not been seen before. We compare our model performance on these writers with a model is biased by the writer and letter identities (the benchmark model). The latter model was not constrained from seeing those writers (thus, the reported results of the comparison overestimates the actual performance of that model). 

\par The BLEU scores can be seen in table \ref{table:bleu_transfer}. Our model achieves on BLEU-3 score 32.2\% and 42.1\% accuracy on the Speed and Freeman code features, compared to 25.3\% and 27.7\% on the benchmark model for the same features respectively. 
\par The EoS analysis can be seen in table \ref{table:EoS_transfer}. Our model achieves a coefficient value of 0.99, compared to 0.5 for the benchmark.
Thus, the new model clearly outperform the current benchmarks on the transfer task, on both BLEU score and EoS analysis.

\subsection{Styles per letters}
\par One of the nice consequences of using our model is that we can have a better look at the styles. We explore the latent space for multiple letters, and see that we can uncover interesting writing styles. A full scale analysis is beyond the scope of this paper. We project the latent space using \textit{Principal Components Analysis} (PCA) \cite{jolliffe2011principal} and t-SNE \cite{maaten2008visualizing}.

\par
\begin{figure}[htbp!]
    \centering
    \fbox{\includegraphics[scale=0.34]{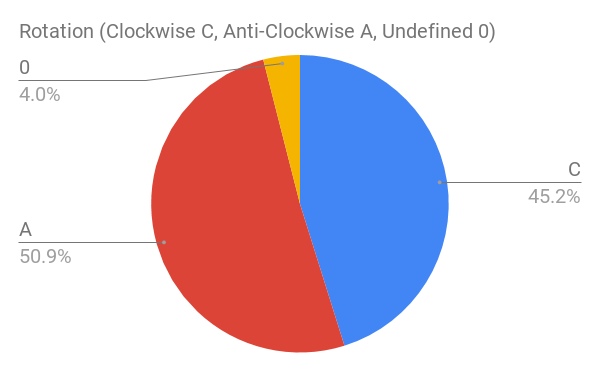}}
    \caption{Results of the manual annotation for the rotation of letter X drawings over the whole dataset. Almost half the writers drew X clockwise, the other half anti-clockwise. The undefined styles were unclear to determine.}
    \label{fig:x_rotation}
\end{figure}

\begin{figure}[htbp!]
    \centering
    \fbox{\includegraphics[scale=0.25]{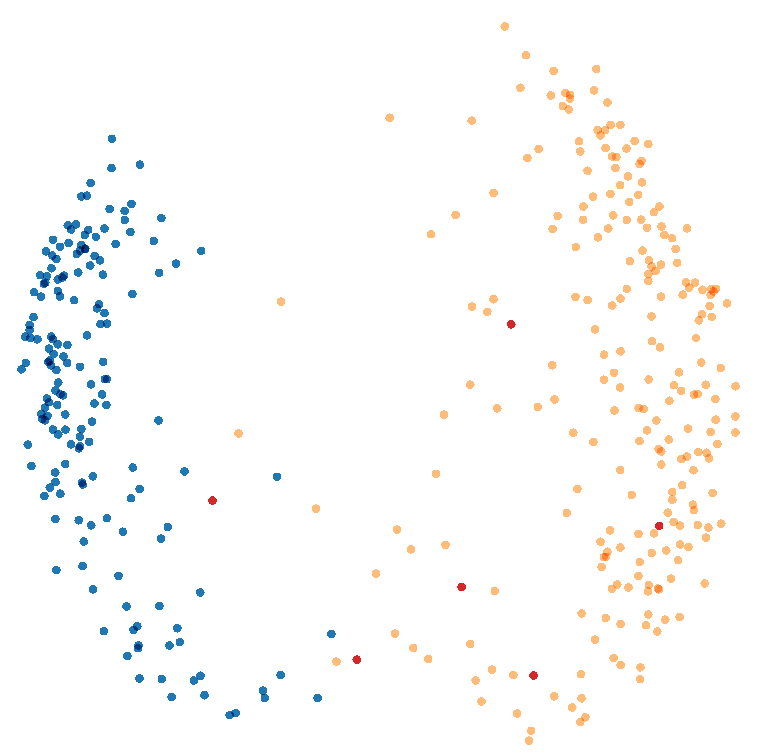}}
    \caption{Projection for latent space for letter X using PCA. The colors show the ground truth of the X rotation: blue is counter clockwise, orange is clockwise, and the few red points are undefined.}
    \label{fig:x_bottleneck}
\end{figure}

\begin{figure}[!htbp]
    \centering
    \begin{subfigure}{0.22\textwidth}
        \includegraphics[scale=0.25]{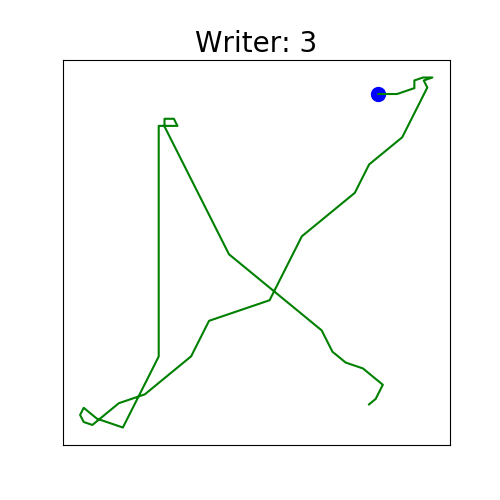}
    \end{subfigure}
    \hspace{0.5em}
    \begin{subfigure}{0.22\textwidth}
        \includegraphics[scale=0.25]{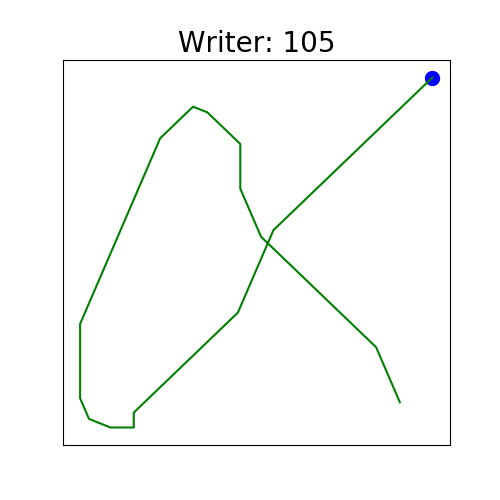}
    \end{subfigure}
    \vspace{1em}
    \begin{subfigure}{0.22\textwidth}
        \includegraphics[scale=0.25]{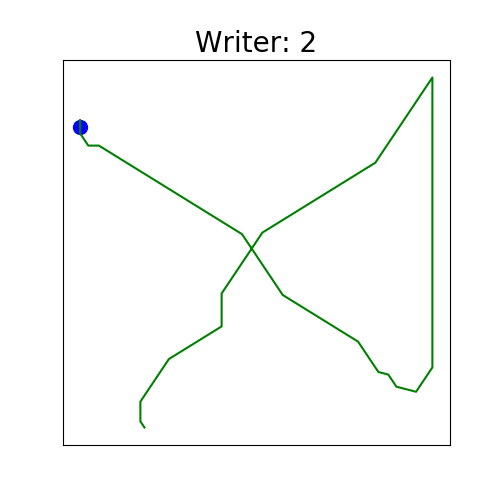}
    \end{subfigure}
    \hspace{0.5em}
    \begin{subfigure}{0.22\textwidth}
        \includegraphics[scale=0.25]{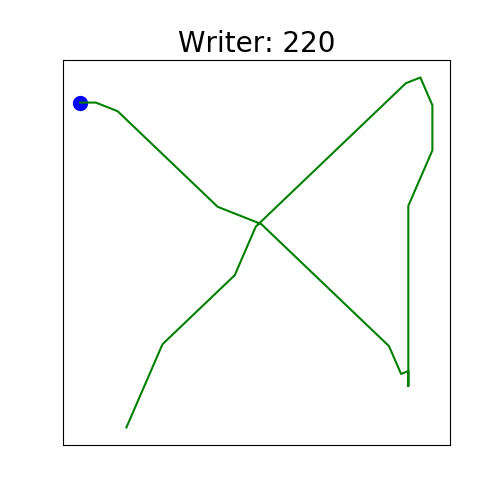}
    \end{subfigure}
    
    \caption{Examples for writing of letter X. Starting point is marked with the blue mark. Each raw is randomly sampled from each cluster in the bottleneck. The clusters shows that almost half the writers draw the letter clockwise (first row, first cluster), and the other half draw it anti-clockwise (second row, second cluster).}
    \label{fig:examples_x}
\end{figure}

 As a start, we take a look at letter X. Beforehand, we identified a style feature in letter X: some writer draw X clockwise, and some draw it anti-clockwise. We manually annotated the whole dataset for this feature; the result can be seen in figure \ref{fig:x_rotation}. Almost half of the writers draw the letter X clockwise, and the other half draw it anti-clockwise. If our assumption is correct, our model should be able to capture this feature. We project the latent  of the model using PCA on all the letter X, which can be seen in figure \ref{fig:x_bottleneck}. The model latent space clusters almost perfectly based on rotation. Examples for letters from both clusters are in figure \ref{fig:examples_x}. 

\par
\begin{figure}[htbp!]
    \centering
    \fbox{\includegraphics[scale=0.2]{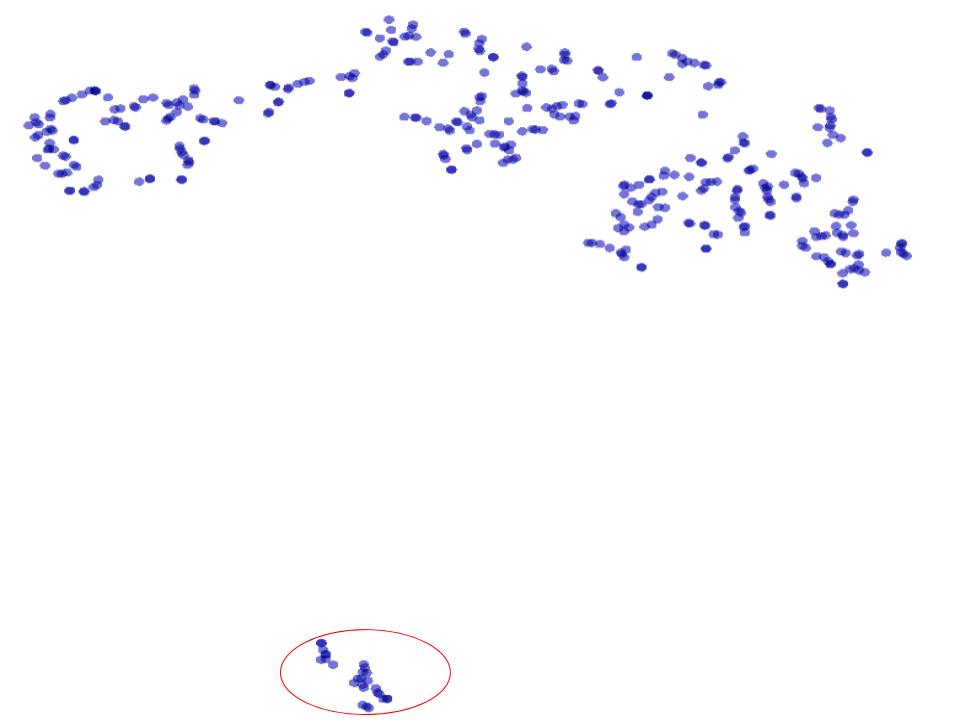}}
    \caption{Projection for latent space for letter C using t-SNE. The cluster surrounded by the red circle has a clear interpretation, where writers have a cursive style.}
    \label{fig:c_letter}
\end{figure}

\begin{figure}[!htbp]
    \centering
    \begin{subfigure}{0.22\textwidth}
        \includegraphics[scale=0.25]{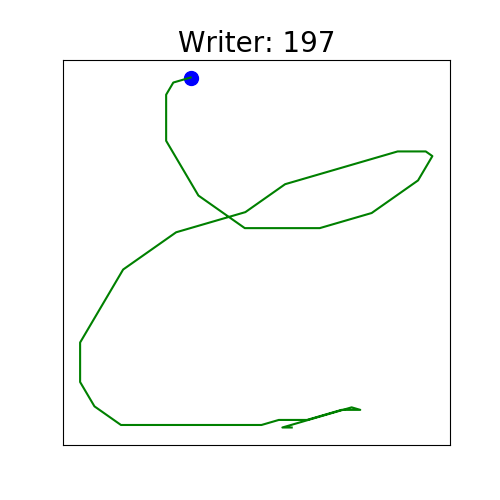}
    \end{subfigure}
    ~
    \begin{subfigure}{0.22\textwidth}
        \includegraphics[scale=0.25]{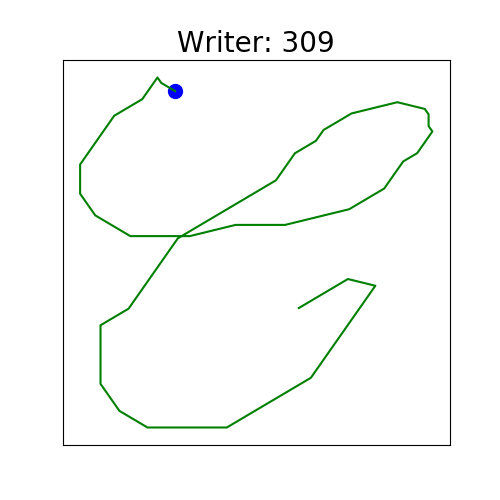}
    \end{subfigure}
    ~
    \begin{subfigure}{0.22\textwidth}
        \includegraphics[scale=0.25]{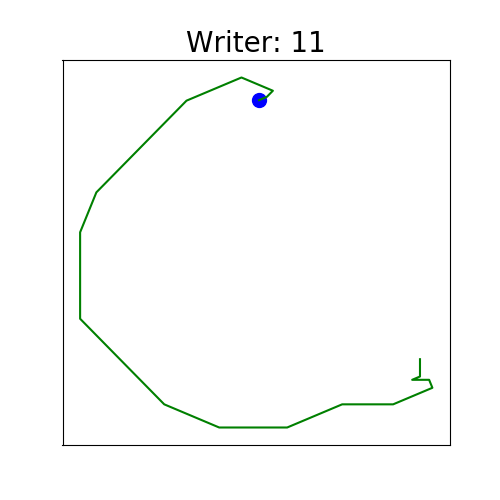}
    \end{subfigure}
    ~
    \begin{subfigure}{0.22\textwidth}
        \includegraphics[scale=0.25]{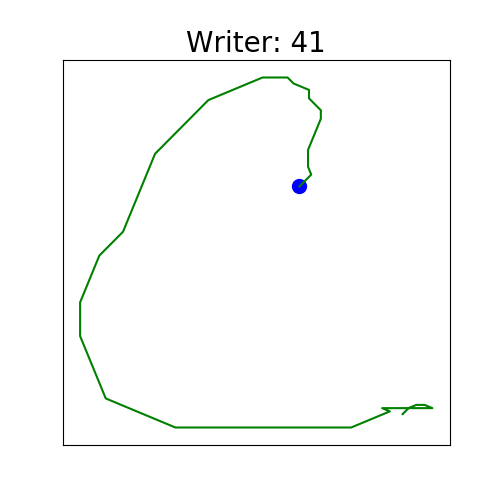}
    \end{subfigure}
    
    \caption{Examples for writing of letter C from the selected cluster (first row) versus the rest of the letter drawings (second row). Starting point is marked with the blue mark. The drawings from the selected cluster show people with Edwardian style of handwriting.}
    \label{fig:examples_c}
\end{figure}

 Encouraged by the results on letter X, we explored more letters. For letter C, we can see the latent space project in figure \ref{fig:c_letter}. It can be seen that there are at least two main clusters. Examples from this cluster in the red ellipse are in figure \ref{fig:examples_c}. The indicated cluster represents the Edwardian handwriting style. The rest of the writers (in the big cluster) have a very similar style (this is expected, since the drawing of the letter C is quite simple).

\par
\begin{figure}[htbp!]
    \centering
    \fbox{\includegraphics[scale=0.20]{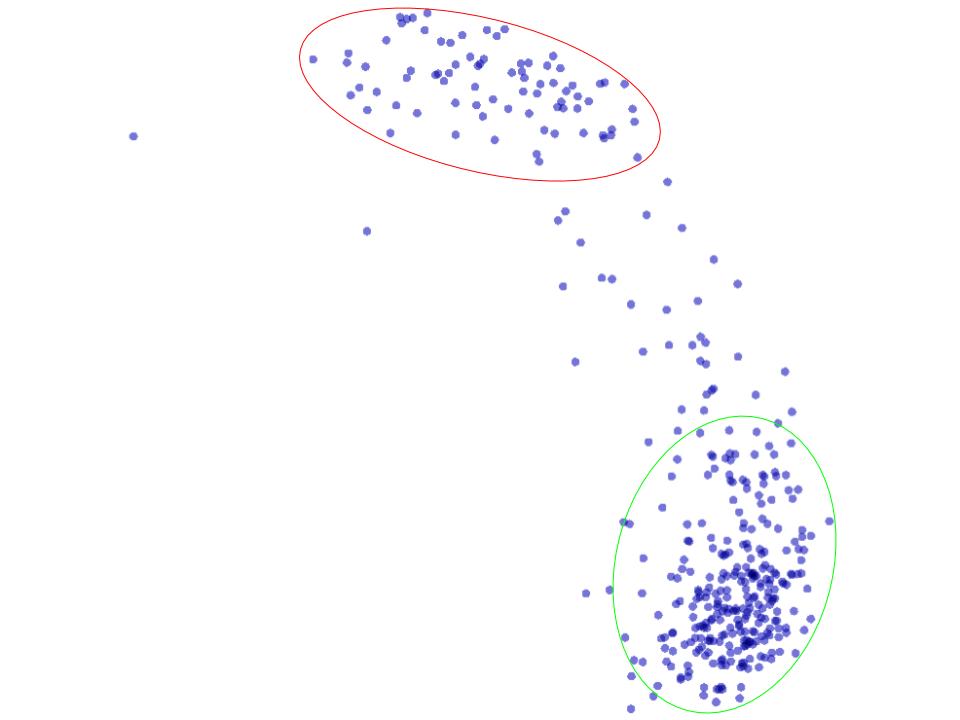}}
    \caption{Projection for latent space for letter A using PCA.}
    \label{fig:a_bottleneck}
\end{figure}

\begin{figure}[!htbp]
    \centering
    \begin{subfigure}{0.22\textwidth}
        \includegraphics[scale=0.25]{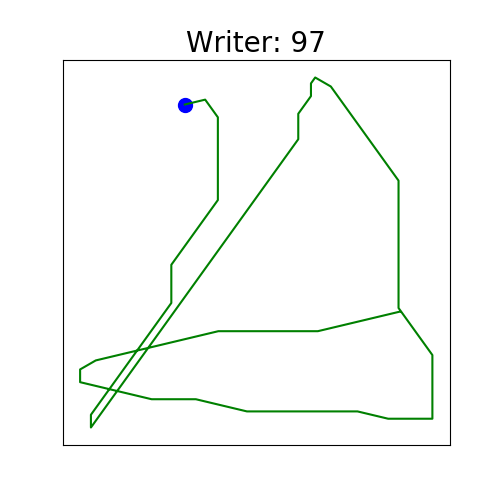}
    \end{subfigure}
    ~
    \begin{subfigure}{0.22\textwidth}
        \includegraphics[scale=0.25]{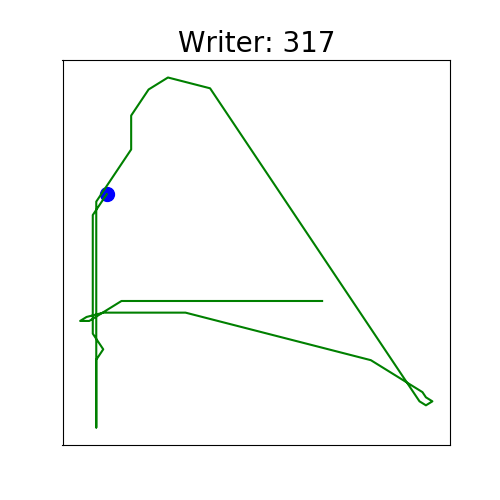}
    \end{subfigure}
    ~
    \begin{subfigure}{0.22\textwidth}
        \includegraphics[scale=0.25]{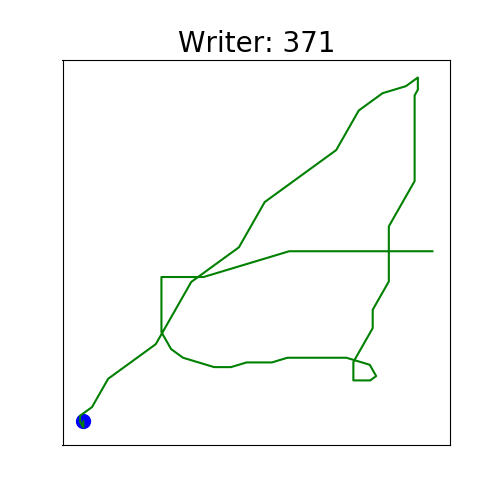}
    \end{subfigure}
    ~
    \begin{subfigure}{0.22\textwidth}
        \includegraphics[scale=0.25]{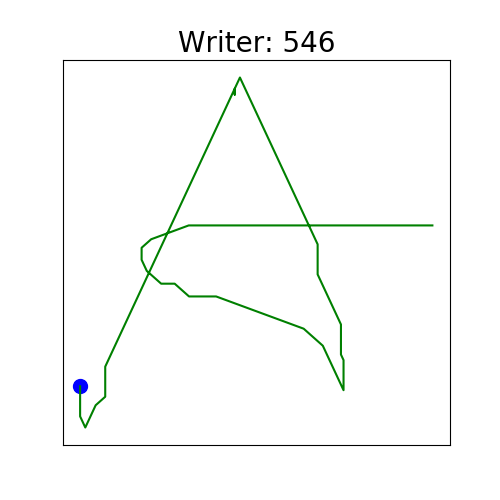}
    \end{subfigure}
    
    \caption{Examples for writing of letter A from the selected clusters. Starting point is marked with the blue mark. Each row is from one cluster. The first row show people who start drawing the letter from the top, going down, and then continue the drawing of the letter. The second row show people who start drawing from down directly. }
    \label{fig:examples_a}
\end{figure}

 For letter A, our model latent space create two main clusters, figure \ref{fig:a_bottleneck}. We give examples from those two in figure \ref{fig:examples_a}, where we can see clear difference in the style. Some people start drawing the letter from down-left, other writers start from the top of letter A, move down, then continue drawing of the letter.

\par
\begin{figure}[htbp!]
\centering
\fbox{\includegraphics[scale=0.20]{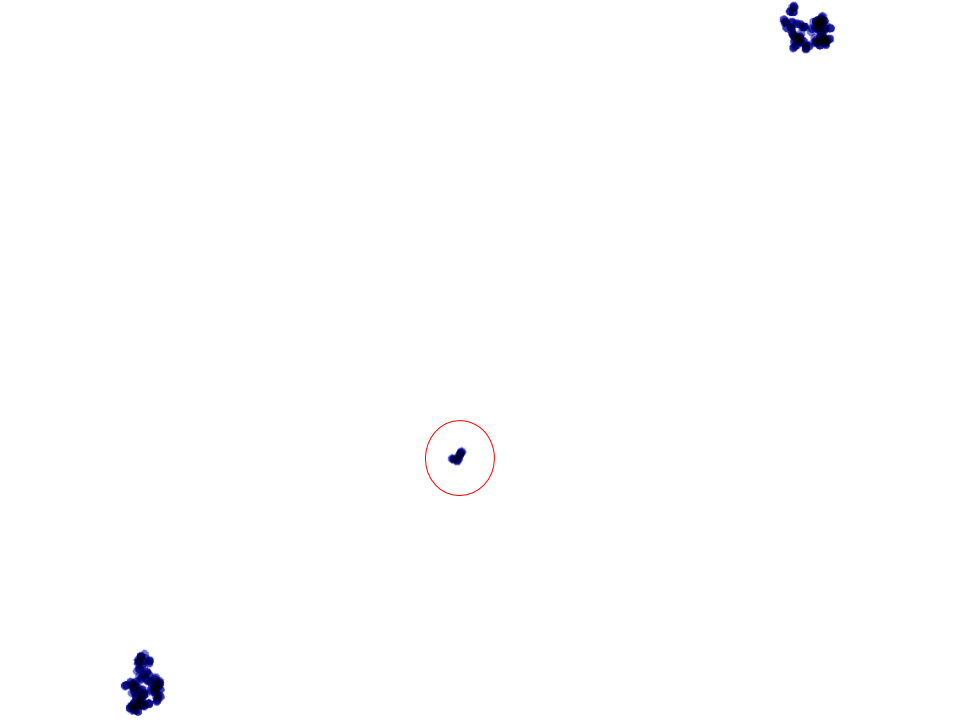}}
\caption{Projection for latent space for letter S using t-SNE. We manage to interpret the indicated cluster as the Edwardian style in drawing. The other two clusters (not indicated) did not show clear difference in the style, but this is an expected behavior from using the t-SNE algorithm, since it does not try to cluster styles as an objective.}
\label{fig:s_bottleneck}
\end{figure}

\begin{figure}[!htbp]
    \centering
    \begin{subfigure}{0.22\textwidth}
        \includegraphics[scale=0.25]{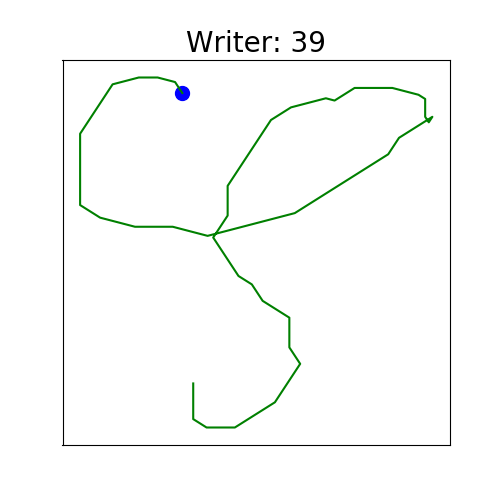}
    \end{subfigure}
    ~
    \begin{subfigure}{0.22\textwidth}
        \includegraphics[scale=0.25]{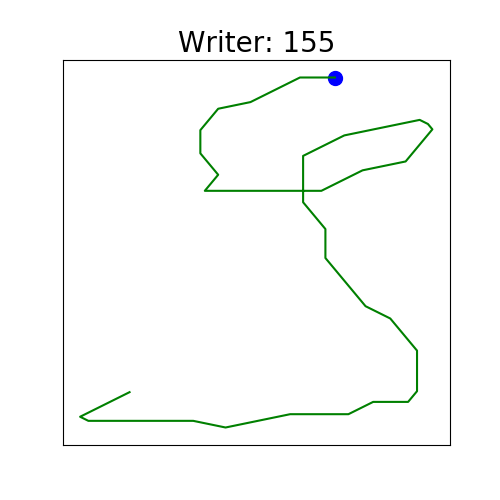}
    \end{subfigure}
    ~
    \begin{subfigure}{0.22\textwidth}
        \includegraphics[scale=0.25]{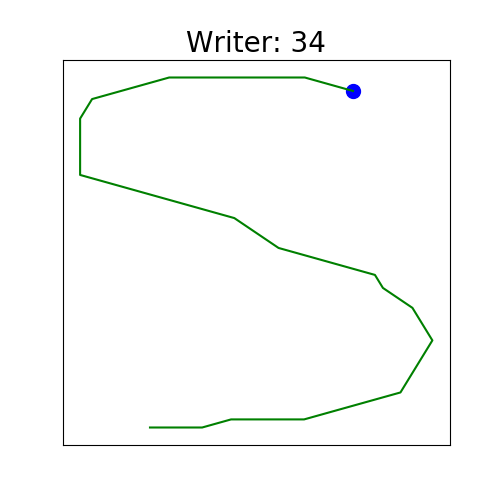}
    \end{subfigure}
    ~
    \begin{subfigure}{0.22\textwidth}
        \includegraphics[scale=0.25]{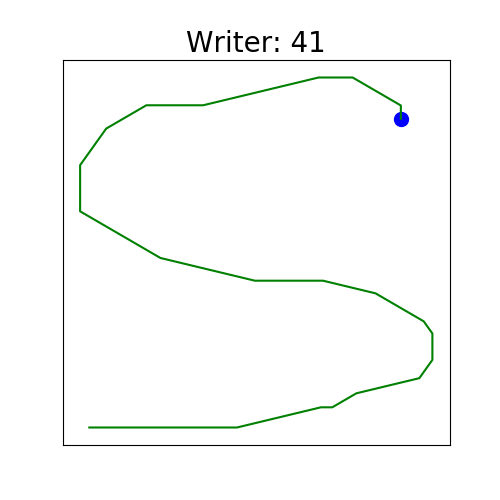}
    \end{subfigure}
    
    \caption{Examples for writing of letter S from the selected cluster (first row) versus the other two clusters (second row). Starting point is marked with the blue mark. The drawings from the selected cluster is always Edwardian style.}
    \label{fig:examples_s}
\end{figure}

 Another example is for letter S bottleneck, figure \ref{fig:s_bottleneck}. There are three resulting clusters which we investigated. The indicated cluster (in red) is clearly different from the other two clusters (not indicated). Examples can be seen in figure \ref{fig:examples_s}. The indicated cluster is again for people with Edwardian handwriting style. We did not find a clear difference between the other two clusters though, but this is an expected outcome of using t-SNE (since it does not have the clear objective of clustering styles).

\par These examples show is that we can use our model to extract verbose style information. 

\section{Conclusions}
\par In this paper, we explored the concepts of styles of handwriting, using a deep  neural network paradigm. We have approached the problem systematically. First, we compared our generation results to the benchmark reported in the state-of-the-art on this problem, and we show that our model outperforms the benchmark. Second, we explore the ability to perform style transfer, by testing the model's performance on 30 new writers. We hypothesize that there is a limited number of style components that describe handwriting, and a good style extraction model should generalize well to new writers. Last, we analyze the latent space of our model for multiple letters, and show that the model separate the different styles in different clusters. 



\section{Future Work}
\par Based on the results of the latent space analysis, our next objective is to build an latent space structure and objective function that disentangle the style manifold. So far, we used multiple projection techniques in order to explore the style information in the latent space. We would like this to emerge on its own in the latent space. This step is usually known as \textit{Knowledge Restructuring}, which enable the addressing of several interesting questions, like:
\begin{itemize}
    \item What are all the different styles available for different letters?
    \item Can we use the styles from those different letters to build a footprint for each writer (i.e. style embedding for the writer)? If so, how good is this embedding in learning to generate letters using it as a prior knowledge only?
    \item If we have a discrete number of styles for each letter, we investigate whether we can predict a writer's style on one letter given the other letters, and what is the contribution of the other letters in identify the style of the writer.
\end{itemize}

\par Also, in this study, we focused only on the upper case letters. We intend to expand our evaluation to include the rest of the dataset (lowercase and digits).

\bibliographystyle{ieeetr}
{\bibliography{bibliography}}

\end{document}